\documentclass{article}

\PassOptionsToPackage{numbers, sort}{natbib}

 \usepackage[main, final]{neurips_2026}

\usepackage[utf8]{inputenc}
\usepackage[T1]{fontenc}
\usepackage{hyperref}
\usepackage{url}
\usepackage{booktabs}
\usepackage{amsfonts}
\usepackage{amsmath}
\usepackage{amssymb}
\usepackage{nicefrac}
\usepackage{microtype}
\usepackage{xcolor}
\usepackage{graphicx}
\usepackage{multirow}
\usepackage{array}
\usepackage{caption}
\usepackage{subcaption}
\usepackage{enumitem}
\usepackage{algorithm}
\usepackage{algpseudocode}

\newcommand{\dtask}{D_{\mathrm{T}}}
\newcommand{\denv}{D_{\mathrm{E}}}
\newcommand{\dtraj}{D_{\tau}}
\newcommand{\dreward}{D_{\mathrm{R}}}
\newcommand{\dW}{D_{\mathrm{W}}}
\newcommand{\atask}{\alpha_{\mathrm{T}}}
\newcommand{\aenv}{\alpha_{\mathrm{E}}}
\newcommand{\atraj}{\alpha_{\tau}}
\newcommand{\aW}{\alpha_{\mathrm{W}}}
\newcommand{\aV}{\alpha_{\mathrm{V}}}
\newcommand{\Linf}{L_{\infty}}

\title{How Data Scales in Agentic Reinforcement Learning: Laws and Synthesis Strategies}

\author{Bowei He$^{1,2}$, Yankai Chen$^{1,2}$, Xiaokun Zhang$^{3}$, Changjiang Han$^1$, Ye Yuan$^2$,\\ \textbf{Chong Li}$^4$, \textbf{Xue Liu}$^{1,2}$\\
  $^1$MBZUAI, $^2$McGill University, $^3$CityUHK, $^4$Columbia University\\
  \texttt{Bowei.He@mbzuai.ac.ae} \\
}

\begin{document}

\maketitle

\begin{abstract}
Pretraining scaling laws treat training data as a one-dimensional quantity: a token count. Agentic reinforcement learning (RL) inherits this scalar abstraction, but modern synthesis pipelines now generate \textit{tasks}, \textit{environments}, and \textit{trajectories} independently and at very different costs, turning data into a structured design space. This raises a question pretraining never had to answer: given a synthesis budget, \textit{which axis} should be scaled? We address this question end-to-end with three tightly coupled contributions. First, we decompose agentic data into three cardinality axes (task, environment, trajectory), plus a reward-density control variable, and fit per-axis scaling relations on different large language models (LLMs) across math+code, SWE, web, and UI/tool application domains. Across all model sizes, model families, domains, and RL algorithms (PPO and GRPO) we measured, the scaling exponents order as environment $>$ task $>$ trajectory. When environment diversity is decomposed into world and environment-structure diversity, its advantage is mainly driven by the former, while the latter is just comparable to the task diversity. Second, we use the fitted laws as a common ruler to price synthesis methods, capturing each method's efficiency as a per-axis discount factor and identifying critical synthetic-to-real ratios at which collapse begins; these ratios differ by an order of magnitude across axes. Third, we cast budget allocation as a constrained optimization under our laws and discounts, deriving a Chinchilla-style compute-optimal recipe and validating it at held-out scale. The empirical results demonstrate that following the recipe yields up to $1.7\times$ compute efficiency over the dominant ``scale-the-trajectories'' practice.
\end{abstract}

\section{Introduction}
\label{sec:intro}
Existing compute-optimal scaling laws for language model training typically reduce the training data distribution to a single scalar, the number of tokens, as exemplified by the Chinchilla scaling law \citep{hoffmann2022chinchilla} and its predecessor framework \citep{kaplan2020scaling}. Recent agentic reinforcement learning (RL) works~\citep{guo2025deepseekr1,mai2025agentrl,khatri2025art} inherited this single-axis abstraction, plotting performance against rollout tokens, prompt count, or training steps as if they were interchangeable.
But agentic data is fundamentally different.
A single ``data point'' simultaneously encodes a task specification, an environment with its API surface and dynamics, a sampled trajectory of model actions and tool feedback, and a reward signal, each generated by a different pipeline at a different cost.
Synthesis pipelines amplify this heterogeneity: environment generators~\citep{wang2026agent,chen2025scaling,guo2025genenv}, task generators~\citep{ramrakhya2025scaling,zhou2025sca}, and trajectory generators~\citep{goldie2025synthetic,chen2026beyond} act on different axes and saturate at different rates.
The scalar abstraction has quietly become wrong.

\begin{figure}[t]
\centering
\includegraphics[width=0.92\textwidth]{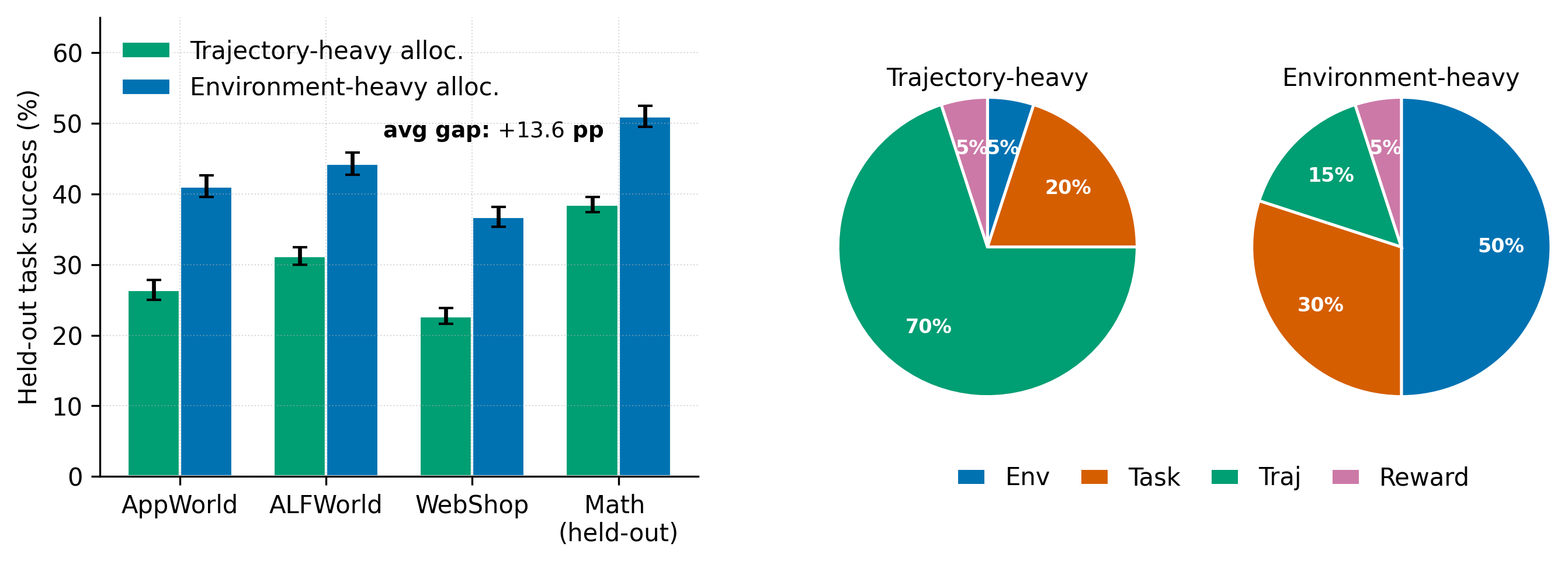}
\vspace{-1mm}
\caption{\textbf{Same compute, very different outcomes.}
Two GRPO runs on Qwen3-8B with identical total rollout tokens and gradient FLOPs, differing only in how the data budget is split across the task, environment, and trajectory axes and the reward control.
\textbf{Left:} task success rate on AppWorld, ALFWorld, WebShop, and a held-out math suite.
\textbf{Right:} the corresponding allocation pies.}
\label{fig:motivating}
\end{figure}

This shift even changes what the central question is. With modern synthesis tooling, agentic data is effectively unbounded in supply, so the practitioner's problem is no longer ``how much more data should I collect,'' but ``where should I spend my synthesis budget?''
Figure~\ref{fig:motivating} makes this concrete: two training runs on Qwen3-8B with identical total tokens and FLOPs, differing only in how the budget is split across (task, environment, trajectory), reach success rates that differ by 14 percentage points.
A single-axis power law cannot fit both, which means the bottleneck has moved from quantity to allocation. However, allocation in current agentic RL pipelines is largely decided by convention or convenience, not by any quantitative account of what each axis is worth.

A principled answer to ``where to synthesize'' requires three conceptual moves, each enabling the next:
(i) \textit{decompose} the data axis into independent design dimensions and measure each independently;
(ii) \textit{price} synthesis methods by how efficiently they convert compute into ``real-equivalent'' data on each dimension;
(iii) \textit{solve} budget allocation as a constrained optimization under the fitted laws and discounts. The order is not optional: without per-axis exponents there is no common unit in which to price synthesis, and without prices there is no objective for the optimization to minimize.

Concretely, we argue that in agentic RL, synthesis turns the data axis from a scalar into a design space, and the right scaling laws turn that design space into a solvable optimization problem. We instantiate this argument with three contributions, each enabling the next:
\begin{itemize}[leftmargin=10pt]
\item \emph{Decomposing the design space (Section \ref{sec:laws}).} We separate agentic data into three cardinality axes, task, environment, and trajectory, plus a reward-density control variable, and fit independent scaling exponents on each axis. Across Qwen3-4B/8B/14B, GRPO and PPO, five domains, and two further model families (twenty one settings in total), we observe a robust ordering: environment $\aenv >$ task $\atask >$ trajectory $\atraj$; resolving the environment axis into world and environment-structure diversity sharpens it to world $\aW >$ environment structure $\aV \approx$ task $\atask >$ trajectory $\atraj$.
\item \emph{Pricing synthesis along each axis (Section \ref{sec:pricing}).} Using the fitted laws as a common ruler, we measure how efficiently existing synthesis methods convert compute into real-equivalent data, capturing this as a per-axis discount $\kappa \in [0,1]$ and identifying critical synthetic-to-real ratios $r^{\star}$ at which collapse begins. We find $r^{\star}_{\mathrm{E}} \approx 9.4 \gg r^{\star}_{\mathrm{T}} \approx 3.4 \gg r^{\star}_{\tau} \approx 1.4$, an ordering we link mechanistically to how grounded each axis is to external truth.
\item \emph{Solving the allocation problem (Section \ref{sec:allocation}).} With laws and discounts in hand, ``what should I synthesize'' becomes a constrained optimization with a tractable solution.
We derive a compute-optimal recipe that specifies how each data-axis budget $D$ and model size $N$ should scale with the total compute budget $C$: $\denv^{\star} \propto C^{0.34}$, $\dtask^{\star} \propto C^{0.27}$, $\dtraj^{\star} \propto C^{0.13}$, $N^{\star} \propto C^{0.26}$, and validate the predicted allocation at a held-out compute scale. Whether the budget is allocated to generating data or building new worlds, trajectory resampling ranks last.
\end{itemize}
Together, these contributions reframe agentic data scaling: the bottleneck is not data quantity but data allocation across a structured design space.

\vspace{-1mm}
 
\section{Related Work}
\label{sec:related}
\textbf{Data scaling laws.} In LLM pretraining, Kaplan et al.~\citep{kaplan2020scaling} and Hoffmann et al.~\citep{hoffmann2022chinchilla} established power-law relationships between loss, parameters, and tokens, giving rise to compute-optimal training; subsequent work has refined these laws under inference constraints~\citep{sardana2023beyond}, robustness checks~\citep{pearce2024reconciling}, and parameter-counting interpretations~\citep{schaeffer2025evaluating}. In RL for LLMs, DeepSeek-R1~\citep{guo2025deepseekr1} demonstrated that pure RL with verifiable rewards scales reasoning capability, ScaleRL~\citep{khatri2025art} characterized RL compute scaling under a fixed recipe, Mai et al.~\citep{mai2025agentrl} reported a ZeroTIR-specific ``Agent RL Scaling Law'', and Tan et al.~\citep{tan2025scaling} provided an empirical study of RL post-training scaling. In both settings, data enters as a single axis: tokens in pretraining, and rollout tokens or unique prompts in RL. Rather than adding a new global axis, we decompose the data axis itself into task, environment, and trajectory axes, and find that this decomposition is necessary to predict allocation outcomes.
 
\textbf{Agentic data synthesis.}
Recent work synthesizes individual axes of agentic data: environments~\citep{wang2026agent,chen2025scaling,guo2025genenv,zeng2025rlve}, tasks~\citep{ramrakhya2025scaling,zhou2025sca,trivedi2024appworld}, and trajectories~\citep{goldie2025synthetic,chen2026beyond,zhang2025agentrl,li2026deepagent}. A growing body of work has documented synthesis-induced quality degradation, or model collapse, in pretraining~\citep{shumailov2023curse,seddikbad,gerstgrasser2024escape,fengbeyond,yi2025escaping}. Existing methods scale a chosen axis in isolation; we provide a framework that prices them on a common scale, finding their collapse ratios differ by an order of magnitude.
 
\textbf{Curriculum and unsupervised environment design.}
GenEnv~\citep{guo2025genenv}, self-evolving curricula~\citep{chen2025selfevolving}, and difficulty-aligned task generation~\citep{ramrakhya2025scaling,zeng2025rlve} address a question our framework deliberately leaves open: \emph{in what order} should data along a given axis be presented? Our decomposition operates one level up, asking how much budget each axis deserves; curriculum operates within an axis, deciding which slice to sample at each training step. The two are complementary: our recipe specifies the per-axis budget, and curriculum methods can then schedule consumption of that budget. 
 
\section{Preliminaries and Setup}
\label{sec:setup}
\subsection{Agentic RL and the data tensor} 
\label{sec:axes}
A policy $\pi_\theta$ acts in an environment instance $e \in \mathcal{S}_E$ on a task $t \in \mathcal{S}_T$,
producing a trajectory $\tau = (o_0, a_0, o_1, a_1, \ldots, o_H)$, where $a_i$ are tool or action calls and $o_i$ are observations, including environment feedback. A reward function $\rho$ assigns $R(\tau,t,e) = \rho(\tau,t,e)$. We define the training corpus as $\mathcal{D} = \bigl(
\mathcal{S}_T, \mathcal{S}_E, \{(t_i,e_i,\tau_i)\}_{i=1}^{n}, \rho
\bigr)$, where $\mathcal{S}_T$ is the finite set of training tasks,
$\mathcal{S}_E$ is the finite set of training environment instances,
and each trajectory $\tau_i$ is associated with a task-environment
pair $(t_i,e_i)$. For later use, let $n_{t,e} = \bigl|\{\,i : t_i=t,\ e_i=e\,\}\bigr|, \Omega = \{(t,e): n_{t,e}>0\}$, where $n_{t,e}$ is the number of trajectories collected for cell $(t,e)$ and $\Omega$ is the set of covered task-environment cells.

The data tensor induces three cardinality axes and one control variable. \textbf{(1) Task axis $\dtask$:} the number of unique training task specifications,
$\dtask = |\mathcal{S}_T|$. Task specifications are deduplicated by sentence-embedding cosine distance $>0.15$; five alternative rules, including embedding-free ones, keep $\atask$ within $[0.238, 0.261]$ (Appendix~\ref{app:dedup}). \textbf{(2) Environment axis $\denv$:}
the number of distinct training environment instances, $\denv = |\mathcal{S}_E|$. An environment-structure instance is identified by its structural signature, a canonical hash over the typed tool/API surface, the state schema, the transition semantics, and the verifier interface; content never changes it, so a procedural seed creates a new instance only if it changes the signature. The signature is a deduplication criterion, not a claim that two instances with distinct signatures constitute distinct worlds; variation in the underlying world, meaning a distinct repository, corpus, app, or site, is reported separately as world diversity $\dW$ (Table~\ref{tab:levels}; Appendix~\ref{app:env}). \textbf{(3) Trajectory axis $\dtraj$:} the average number of rollouts per covered task-environment cell, $\dtraj = \frac{1}{|\Omega|} \sum_{(t,e)\in\Omega} n_{t,e}$. Increasing $\dtraj$ corresponds to repeated sampling within already covered task-environment cells, which is the standard practice in GRPO~\citep{shao2024deepseekmath}. In the balanced full-factorial case, where every $(t,e)\in\mathcal{S}_T\times\mathcal{S}_E$ has the same number of rollouts, this reduces to $\dtraj = \frac{n}{\dtask \denv}$. \textbf{(4) Reward control variable $\dreward$:} the information density of the reward signal, used as a categorical sparse/dense control rather than a cardinality axis: sparse rewards provide only an outcome-level signal, whereas dense rewards provide at least three verifiable intermediate signals per trajectory (Appendix~\ref{app:reward}).

\begin{table}[t]
\centering
\caption{\textbf{Three levels of environment variation} (Section~\ref{sec:world}). Only L1 and L2 create new instances.}
\label{tab:levels}
\footnotesize
\setlength{\tabcolsep}{4pt}
\begin{tabular}{@{}>{\raggedright\arraybackslash}p{0.13\linewidth}>{\raggedright\arraybackslash}p{0.29\linewidth}>{\raggedright\arraybackslash}p{0.31\linewidth}>{\raggedright\arraybackslash}p{0.21\linewidth}@{}}
\toprule
Level & What varies & Example & Exponent \\
\midrule
L0: content & seed or initial state within the same repository and harness & a new AIME problem; a new issue on the same repository & $0.07\pm0.02$ (placebo) \\
L1: structure & tool/API surface, state schema, transition semantics, verifier & stateless vs.\ persistent REPL; pytest vs.\ tox output grammar & $\aV = 0.28\pm0.03$ ($0.41$ on math+code) \\
L2: world ($\dW$) & repository, corpus, app, or site with its own state & a repository or app unseen in training & $\aW = 0.44\pm0.04$ \\
\bottomrule
\end{tabular}
\end{table}
 
\subsection{Experimental testbeds}
\label{sec:testbeds}
To stress-test the law's domain invariance, we use four domains:
(a)~\textbf{Math+code (TIR)}: Python interpreter as tool, training on synthetic math problems following the ZeroTIR recipe~\citep{li2025torl,mai2025agentrl}, evaluated on AIME24/25~\citep{maa_aime}, MATH500~\citep{hendrycks2measuring}, AMC23~\citep{maa_amc}.
(b)~\textbf{Code agents (SWE)}: SWE-Gym~\citep{pantraining}-style execution environments, evaluated on SWE-bench Verified~\citep{jimenez2024swebench}.
(c)~\textbf{Web agents}: WebArena-Lite~\citep{zhou2024webarena,wei2025webagent} with the BrowserGym~\citep{chezelles2025browsergym} interface.
(d)~\textbf{UI/tool agents}: AppWorld~\citep{trivedi2024appworld} and ALFWorld~\citep{shridhar2021alfworld}.
A fifth domain, (e)~\textbf{search agents} (Wikipedia retrieval plus a paginated web-search API; Bamboogle~\citep{press2023measuring} and a multi-hop split), serves as a robustness check. Evaluation is world-held-out on SWE and AppWorld but task-held-out on web (Appendix~\ref{app:setup}).
For axes that require synthesis, we use AWM~\citep{wang2026agent} for environments, AutoPlay~\citep{ramrakhya2025scaling}/SCA~\citep{zhou2025sca} for tasks, and standard policy-rollout for trajectories.
 
\textbf{Models and training.}
We use Qwen3-4B / 8B / 14B base checkpoints~\citep{yang2025qwen3} as the size sweep, with Qwen3-8B as the default, plus Llama-3.1-8B~\citep{dubey2024llama} and Qwen2.5-7B~\citep{yang2024qwen25}. \textbf{GRPO}~\citep{shao2024deepseekmath} (default; group size 8, no critic) and \textbf{PPO}~\citep{schulman2017ppo} (with a separate critic of equal size; details in Appendix~\ref{app:setup}) are employed as the agentic reinforcement learning algorithms.
All experiments use the verl framework~\citep{sheng2025hybridflow} with VerlTool~\citep{jiang2025verltool} for multi-tool rollout.
We log full per-axis sweeps; all reported numbers are means over 3 seeds with bootstrap 95\% confidence intervals (CIs).
 
\textbf{Loss-equivalent metric.}
Because RL with verifiable rewards has no smooth ``loss,'' we use $L \equiv -\log p_{\mathrm{success}}$ where $p_{\mathrm{success}}$ is task success rate on a held-out evaluation set, following~\citep{khatri2025art,tan2025scaling}.
This makes our laws directly comparable to pretraining laws and allows clean log-linear fits (Appendix~\ref{app:metric}).
\vspace{-4mm} 
\section{Decomposing the Design Space: Per-Axis Scaling Laws}
\label{sec:laws}
\vspace{-2mm} 

Before deciding where to synthesize data, we first quantify the marginal scaling behavior of each axis. This section implements the first of the three steps outlined in Section~\ref{sec:intro}: we decompose agentic data into its three cardinality axes and fit an independent scaling exponent for each.
 
\textbf{Functional form.} To make the three cardinality axes comparable, we need a functional form in which each axis contributes its own scaling term and its own exponent. Following the additive power-law tradition of pretraining scaling laws~\citep{hoffmann2022chinchilla,kaplan2020scaling}, we posit:
\begin{equation}
L(\dtask, \denv, \dtraj) - \Linf
\;=\; A\cdot \dtask^{-\atask} + B\cdot \denv^{-\aenv} + C\cdot \dtraj^{-\atraj}
\;+\; I(\dtask, \denv, \dtraj),
\label{eq:law}
\end{equation}
where $\Linf$ is the irreducible loss and $I$ captures cross-axis interactions that we measure separately in Section~\ref{sec:interaction}.
The independent exponents $\alpha_i$ are the loss-reduction-per-log-unit of each axis, a unit-normalized currency we use throughout the paper to compare and price synthesis methods. We call such fitted per-axis scaling relations laws for brevity; we claim the ordering of their exponents, not their values.
 
\subsection{Independent-axis exponents}
\label{sec:exponents}
\textbf{Protocol.} For each axis, we hold the other two at values fixed in advance by a calibration sweep (doubling an axis changes success by less than 2 points), $(\denv, \dtask, \dtraj) = (256, 10{,}240, 8)$, reused unchanged in every cell (Appendix~\ref{app:setup}), and sweep the target axis over 6 nested grid points spanning $32\times$, extended to 8 points as a stability check. Each sweep is repeated for Qwen3-4B/8B/14B in the math+code domain. The sweeps match training signal: every $\denv$ and $\dtask$ grid point uses the same 256{,}000 task-environment cells (8 rollouts each) and 1{,}000 updates, so rollout tokens drift by under $4\%$ and per-environment repetition even falls $32\times$ along $\denv$; only $\dtraj$ scales tokens ($32\times$), and equalizing tokens or FLOPs lowers $\atraj$ to $0.07$ to $0.08$ while moving $\aenv$ and $\atask$ by at most $0.01$ (Appendix~\ref{app:signal}). We fit $L = K D^{-\alpha} + \Linf$ by nonlinear weighted least squares, profiling $\Linf$ over a grid and solving the inner log-linear problem in closed form (Algorithm~\ref{alg:fit}).
 
\begin{figure}[t]
\centering
\includegraphics[width=0.95\textwidth]{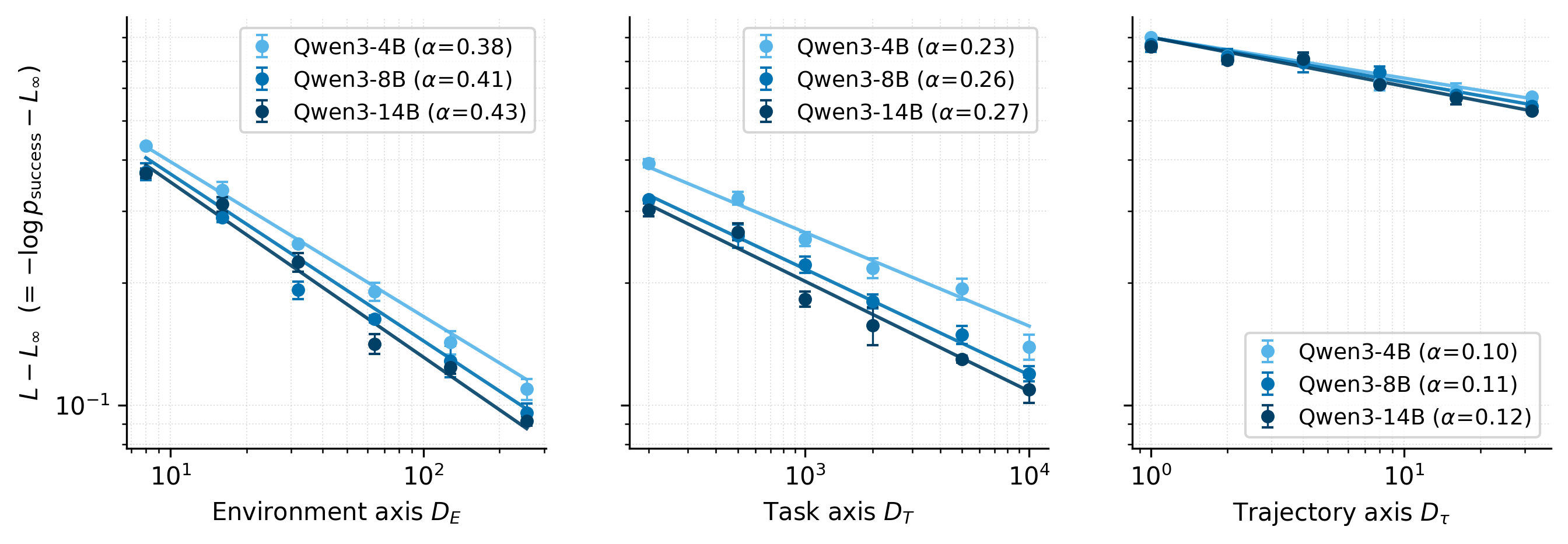}
\vspace{-2mm}
\caption{\textbf{Per-axis scaling laws under controlled sweeps.}
Each panel sweeps one axis with the other two held at saturation; symbols are mean$\pm$SE over 3 seeds, lines are weighted log-log fits.
Across all three model sizes, environment scaling is the steepest, trajectory scaling the flattest.
The exponent ordering $\aenv > \atask > \atraj$ holds with non-overlapping 95\% CIs at every model size.}
\vspace{-2mm}
\label{fig:peraxis}
\end{figure}
 
\textbf{Result.} Figure~\ref{fig:peraxis} shows clean power-law fits ($R^2 > 0.97$ on every panel) with the following fitted exponents (Qwen3-8B, math+code; full table in Appendix~\ref{app:fits}):
\begin{equation}
\aenv = 0.41\pm 0.03,\quad \atask = 0.26\pm 0.02,\quad \atraj = 0.11\pm 0.02.
\label{eq:exponents}
\end{equation}
The ordering $\aenv > \atask > \atraj$ holds with non-overlapping CIs at every model size we tested. It also holds without any fit, as local log-log slopes never overlap across axes (49 of 49 comparisons; Mann-Whitney~\citep{mann1947test} $p = 2.9\times10^{-4}$), and paired bootstrap differences are $\aenv - \atask = 0.148$ $[0.109, 0.187]$ and $\atask - \atraj = 0.150$ $[0.116, 0.184]$ (Appendix~\ref{app:fits}).
 
\textbf{Interpretation.}
Environment diversity is the highest-leverage axis because each new environment forces the policy to generalize across different state spaces, action vocabularies, and tool-feedback dynamics, which are structural variation forms that cannot be obtained from new tasks within a fixed environment; Section~\ref{sec:world} locates most of this leverage in world diversity.
Task diversity is intermediate: more tasks expose new goal structures, but the underlying interaction grammar is fixed.
Trajectory resampling has the lowest exponent because additional rollouts on already-covered (task, env) pairs primarily reduce variance within an already-explored region of policy space; once a sample is informative enough to estimate advantage well, further rollouts give diminishing returns~\citep{shao2024deepseekmath,yu2025dapo}.
 
\subsection{Cross-axis interactions}
\label{sec:interaction}
 \vspace{-3mm}
\begin{figure}[t]
\centering
\includegraphics[width=0.85\textwidth]{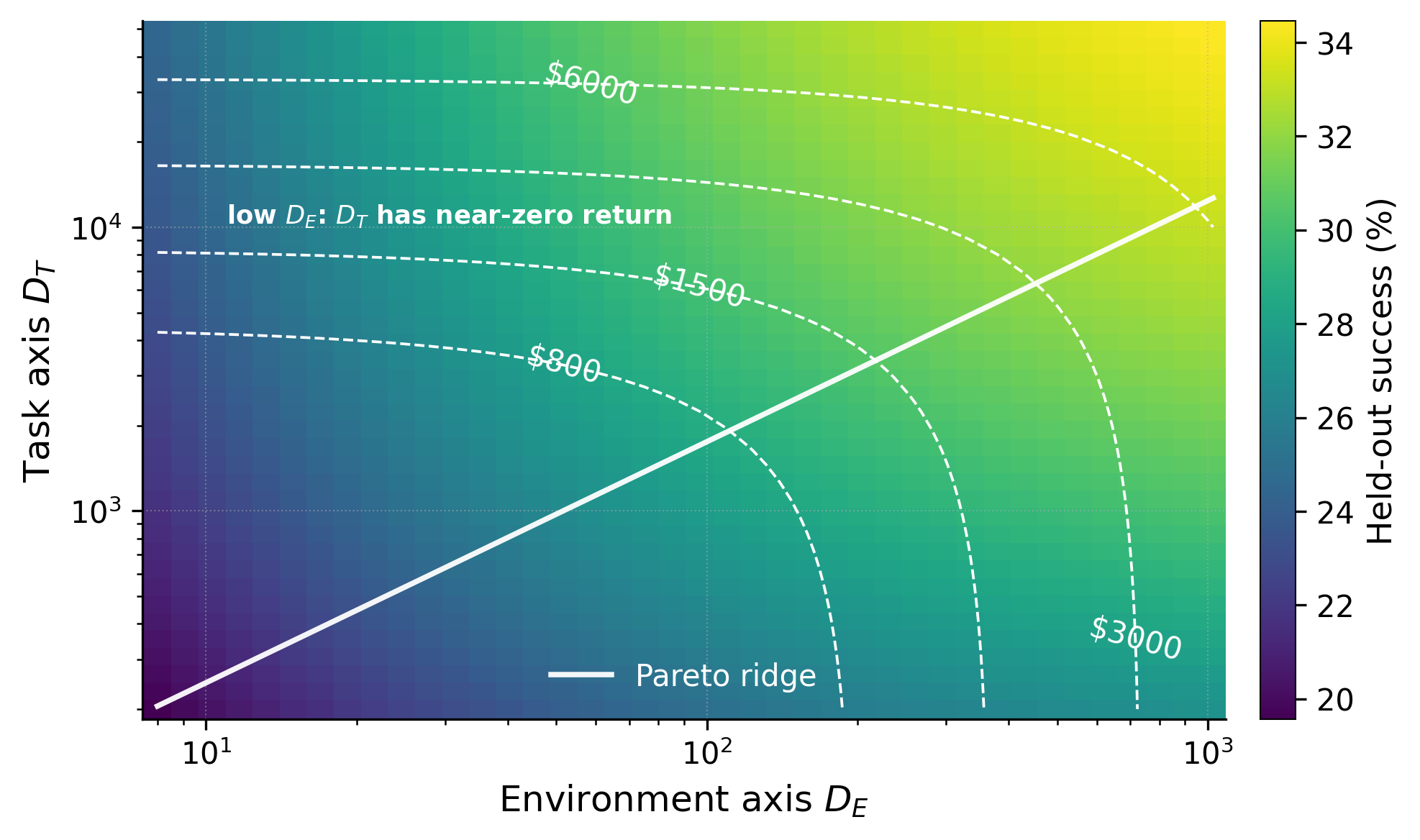}
\vspace{-3mm}
\caption{\textbf{Cross-axis interaction.}
Heatmap of held-out success on Qwen3-8B (math+code) over a $(\denv, \dtask)$ grid at fixed $\dtraj=8$.
Iso-cost contours (dashed) reveal a Pareto ridge: at low $\denv$, scaling $\dtask$ has near-zero return. Environment diversity is a prerequisite for task diversity to matter.}
\label{fig:interaction}
\end{figure}
 
Pure independence (Eq.~\ref{eq:law} with $I\!\equiv\!0$) is a first-order approximation: tasks are bound to compatible environments (9 to 41 per task on average; Appendix~\ref{app:env}), and $I$ captures the residual coupling.
Figure~\ref{fig:interaction} shows that at low $\denv$, scaling $\dtask$ yields almost no gain, because the surface bends sharply rather than being separable.
Fitting a multiplicative correction $I = -\gamma \cdot (\denv\,\dtask)^{-\beta}$ improves $R^2$ from $0.93$ to $0.98$ with $\gamma = 0.18, \beta = 0.21$.
The substantive scientific takeaway is the prerequisite structure: environment diversity gates the value of task diversity.
Practically, this means that pipelines that pour task synthesis into a small environment pool (e.g., many tasks in $\tau$-bench~\citep{barres2025taubench} or its small-scale variants) will see early saturation regardless of how many tasks they generate.
 
\subsection{Robustness across model sizes, algorithms, domains, and model families}
\label{sec:robustness}
 
\begin{table}[t]
\centering
\caption{\textbf{Fitted exponents across model sizes, algorithms, and domains.}
Numbers are mean $\pm$ bootstrap 95\% CI half-width over 3 seeds. Note the ordering $\aenv > \atask > \atraj$ holds in all 18 settings. $^\dagger$No separable world axis: $\aenv$ is a structure exponent (Section~\ref{sec:world}).}
\label{tab:robust}
\small
\begin{tabular}{lll ccc}
\toprule
Model & Algo. & Domain & $\aenv$ & $\atask$ & $\atraj$ \\
\midrule
\multirow{4}{*}{Qwen3-4B}
  & GRPO & Math+Code$^\dagger$  & $0.38\pm0.04$ & $0.23\pm0.03$ & $0.10\pm0.02$ \\
  & GRPO & SWE        & $0.36\pm0.05$ & $0.21\pm0.04$ & $0.09\pm0.03$ \\
  & GRPO & Web        & $0.43\pm0.04$ & $0.26\pm0.03$ & $0.12\pm0.02$ \\
  & PPO  & Math+Code$^\dagger$  & $0.35\pm0.04$ & $0.21\pm0.03$ & $0.10\pm0.02$ \\
\midrule
\multirow{4}{*}{Qwen3-8B}
  & GRPO & Math+Code$^\dagger$  & $0.41\pm0.03$ & $0.26\pm0.02$ & $0.11\pm0.02$ \\
  & GRPO & SWE        & $0.39\pm0.03$ & $0.24\pm0.02$ & $0.10\pm0.02$ \\
  & GRPO & UI/Tool    & $0.45\pm0.03$ & $0.27\pm0.02$ & $0.13\pm0.02$ \\
  & PPO  & Math+Code$^\dagger$  & $0.39\pm0.03$ & $0.24\pm0.02$ & $0.11\pm0.02$ \\
\midrule
\multirow{2}{*}{Qwen3-14B}
  & GRPO & Math+Code$^\dagger$  & $0.43\pm0.03$ & $0.27\pm0.02$ & $0.12\pm0.02$ \\
  & GRPO & Web        & $0.46\pm0.03$ & $0.28\pm0.02$ & $0.13\pm0.02$ \\
\bottomrule
\end{tabular}
\end{table}
\vspace{-2mm}
 
Table~\ref{tab:robust} reports the fitted exponents; across all 18 (model$\times$algo$\times$domain) cells we measured, the ordering $\aenv > \atask > \atraj$ holds with non-overlapping CIs.
Absolute exponents shift modestly: web/UI domains have larger $\aenv$ (more diverse tools amplify the value of environment generalization), and PPO/GRPO yield similar exponents (within 1 SE), suggesting the ordering is an algorithm-invariant property of agentic data, not an artifact of GRPO's variance-reduction mechanics.

\textbf{Other families and a search domain.} Reduced 4-point sweeps preserve the ordering on Llama-3.1-8B $(0.36, 0.23, 0.09)$, Qwen2.5-7B $(0.38, 0.24, 0.10)$, and the search domain $(0.42, 0.25, 0.12)$: it holds in all 21 settings while absolute values shift by up to $0.05$, so the constants are regime-specific (Appendix~\ref{app:fits}).
 
\textbf{Reward control variable.}
Switching from sparse (outcome only) to dense (process-shaped) rewards~\citep{lightman2023let,setlur2024rewarding} on the same data corpus yields a $1.4\times$ to $2.1\times$ effective multiplier on $\dtraj$ but does not change $\aenv$ or $\atask$ (Appendix~\ref{app:reward}). We therefore treat $\dreward$ as a control variable, claim no comparable reward scaling law, and fold its multiplier into an effective trajectory count in Section~\ref{sec:allocation}.

\subsection{Resolving the environment axis: world versus structure}
\label{sec:world}
Outside math+code, our pools grow the number of distinct worlds $W$ (repositories, apps, sites) and of structural variants per world $V$ together, $\denv = W\times V$; the 128 SWE instances, for example, span 11 repositories. Math+code has a single world (a Python tool server), so its $\aenv$ is a pure structure exponent ($^\dagger$ in Table~\ref{tab:robust}). Holding $\denv$, compute, and training signal fixed and moving budget from $V$ to $W$, from $(W, V) = (1, 64)$ to $(8, 8)$ ($(4, 16)$ on web), raises held-out-world success at every step in SWE, UI/tool, and web, with non-overlapping CIs (e.g., from $18.0$ to $26.1\%$ on SWE; Appendix~\ref{app:world}). A joint fit gives $\aW = 0.44\pm0.04$ and $\aV = 0.28\pm0.03$ (a repository-only SWE sweep gives $\aW = 0.46\pm0.05$), so wherever the two separate,
\[
\aW\,(0.44) \;>\; \aV\,(0.28) \;\approx\; \atask\,(0.24 \text{ to } 0.27) \;>\; \atraj\,(0.11).
\]
World diversity is the steepest axis we measured, while structure and task diversity are not separable at our resolution: environment diversity beats task diversity through its world component. The blended $\aenv$ is the composition-weighted mixture of $\aW$ and $\aV$ (on SWE, $0.87\times0.44 + 0.13\times0.28 \approx 0.42$ against the measured $0.39\pm0.03$), and without a world axis, structure carries the whole generalization burden ($\aV = 0.41$ on math+code). Content-only variation is not environment diversity (seeds at a fixed signature give $0.07\pm0.02$), and structural diversity keeps paying beyond factor-level coverage up to an inventory ceiling of 384 signatures (Appendix~\ref{app:world}).

\section{Pricing Synthesis: A Per-Axis Discount Factor}
\label{sec:pricing}
The per-axis exponents fitted in Section~\ref{sec:laws} provide us with a ruler for how data scales. With that ruler in hand, we can finally compare heterogeneous synthesis methods on a common scale, asking how efficiently each one converts compute into real-equivalent data along the axis it targets. 
 
\begin{table}[t]
\centering
\caption{\textbf{Synthesis taxonomy along the data axes.}
Rows: representative synthesis methods, all run through one common harness (Appendix~\ref{app:kappa-est}).
Columns: which axis each primarily scales, the effective discount $\kappa$ measured at $r = D_s/D_r = 1$ (equal real and synthetic), the axis-specific collapse ratio $r^{\star}$, and rough cost per ``unit'' (1k tasks, 1 environment, or 1 trajectory) on Qwen3-8B. The estimator of Section~\ref{sec:collapse} gives $r^\star = 9.4$, $3.4$, $1.4$ for AWM, AutoPlay, SWiRL.}
\label{tab:taxonomy}
\small
\begin{tabular}{l c c c c}
\toprule
Method & Primary axis & $\kappa$@$r{=}1$ & $r^{\star}$ & Cost/unit \\
\midrule
AWM~\citep{wang2026agent}                & $\denv$  & $0.71\pm0.05$ & $9.4\pm0.9$ & 4.10 USD/env \\
GenEnv~\citep{guo2025genenv}        & $\denv$  & $0.66\pm0.06$ & $8.1\pm1.0$ & 3.80 USD/env \\
DreamGym~\citep{chen2025scaling}      & $\denv$\,$+$\,$\dtraj$ & $0.59\pm0.05$ & $6.2\pm0.8$ & 0.04 USD/traj \\
\midrule
AutoPlay~\citep{ramrakhya2025scaling}      & $\dtask$ & $0.54\pm0.04$ & $3.1\pm0.4$ & 0.18 USD/task \\
SCA~\citep{zhou2025sca}              & $\dtask$ & $0.51\pm0.05$ & $2.8\pm0.4$ & 0.21 USD/task \\
RLVE~\citep{zeng2025rlve}       & $\dtask$ & $0.57\pm0.04$ & $3.4\pm0.4$ & 0.15 USD/task \\
\midrule
SWiRL~\citep{goldie2025synthetic}        & $\dtraj$ & $0.34\pm0.06$ & $1.4\pm0.3$ & 0.02 USD/traj \\
TDScaling~\citep{chen2026beyond}   & $\dtraj$ & $0.42\pm0.05$ & $1.7\pm0.3$ & 0.02 USD/traj \\
Self-Play~\citep{liu2025spiral} & $\dtraj$ & $0.31\pm0.06$ & $1.2\pm0.3$ & 0.02 USD/traj \\
\bottomrule
\end{tabular}
\end{table}
 
Table~\ref{tab:taxonomy} maps a representative cross-section of agentic synthesis methods to the axes they primarily scale.
This taxonomy alone organizes a fragmented field by giving every method a shared coordinate system; the $\kappa$ and $r^{\star}$ columns (defined in Sections~\ref{sec:kappa} and~\ref{sec:collapse}) make the comparison quantitative. Reduced to a common generate, filter, and train core, axis membership still explains 97\% of the variance of $\kappa$ (ANOVA $F(2,5) = 76.6$, $p < 0.001$; Appendix~\ref{app:kappa-est}).
 
\subsection{The synthesis discount factor $\kappa$}
\label{sec:kappa}
 
Let $D_r$ and $D_s$ denote real and synthetic data on a target axis.
We posit
\begin{equation}
L(D_r, D_s) \;=\; f\!\big(D_r + \kappa(r)\cdot D_s\big),
\qquad r \equiv D_s / D_r,
\label{eq:kappa}
\end{equation}
where $f$ is the fitted single-axis law and $\kappa(r)\in[0,1]$ is a real-data-equivalent yield.
Crucially, $\kappa$ is meaningful precisely because Section~\ref{sec:laws} provides a conversion rate: an exponent $\alpha$ tells how loss responds to data, so a synthesis method's effective contribution can be measured in real-data units. Without the per-axis exponent, ``synthetic data is X\% as good'' would have no unit.
 
\textbf{Estimation.} For each method and axis, we hold total $D = D_r + D_s$ constant and sweep $r \in \{0, 0.25, 1, 4, 16\}$, fitting $\kappa$ from the resulting performance curve via Eq.~\ref{eq:kappa}; collapse curves use a denser grid.
Errors are bootstrapped over 3 training seeds.

\begin{figure}[t]
\centering
\includegraphics[width=0.82\textwidth]{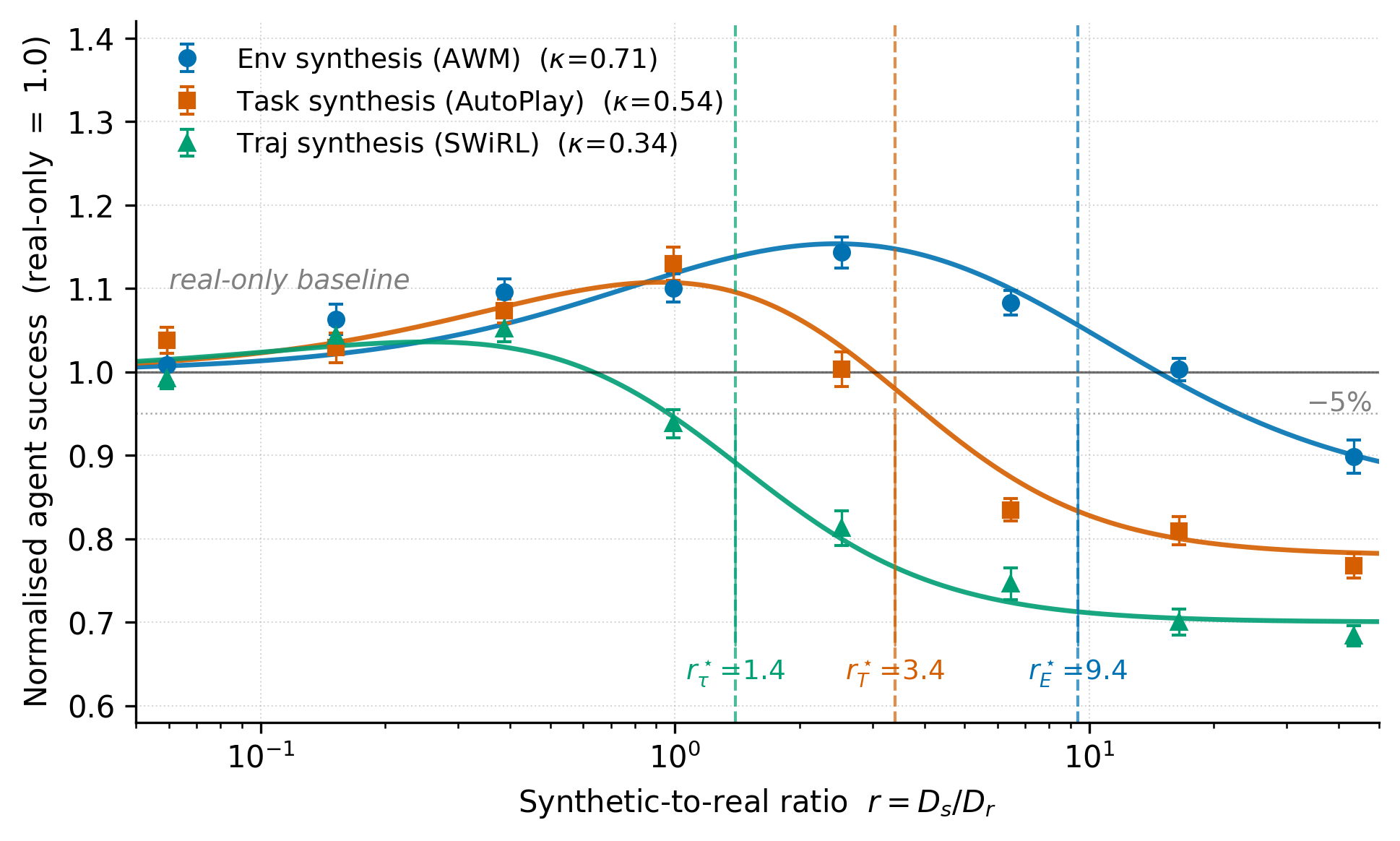}
\vspace{-3mm}
\caption{\textbf{Synthesis collapse, per axis.}
$x$-axis: synthetic-to-real ratio $r=D_s/D_r$ (log scale).
$y$-axis: agent success rate on a held-out evaluation, normalized to the $r=0$ (real-only) value of the same seed group.
Vertical dashed lines mark fitted $r^{\star}$, the ratio at which the curve drops 5\% below the real-only baseline.
Environment synthesis remains net-positive up to $r{\approx}9$, while trajectory synthesis collapses by $r{\approx}1.4$.}
\label{fig:collapse}
\end{figure}
\vspace{-3mm}

\subsection{Per-axis collapse points}
\label{sec:collapse} 
We define the critical collapse ratio $r^{\star}$ as the ratio at which normalized success, past its peak, falls 5\% below the real-only baseline, estimated as the $0.95$ crossing of a monotone logistic fit in $\log r$ on a 13-point ratio grid (Appendix~\ref{app:per-axis-collapse}).
Figure~\ref{fig:collapse} plots normalized success vs.\ $r$ for the three axes (using AWM, AutoPlay, and SWiRL as representatives). Notably, we find
\begin{equation}
r^{\star}_{\mathrm{E}} \approx 9.4 \;\gg\; r^{\star}_{\mathrm{T}} \approx 3.4 \;\gg\; r^{\star}_{\tau} \approx 1.4,
\end{equation}
an order-of-magnitude spread across axes that mirrors the exponent ordering on the same testbed and survives alternative collapse criteria (Appendix~\ref{app:per-axis-collapse}).
 
\textbf{Mechanism: groundedness of the synthesis target.}
The ordering aligns with how strongly each axis is grounded by external truth.
\textbf{Environments} synthesized as executable code are checked at every rollout step by an interpreter, providing a hard constraint that bounds distributional drift; this is consistent with the design principles of AWM~\citep{wang2026agent} and DeepSeek-V3.2's environment pipeline~\citep{liu2025deepseekv3}.
\textbf{Tasks} are anchored less directly. Synthetic tasks are grounded only by whether their reward is verifiable, leaving room for distributional drift in problem complexity.
\textbf{Trajectories} synthesized by sampling from the policy itself~\citep{goldie2025synthetic} are grounded only by the model's own distribution, reproducing and amplifying its biases; this is the agentic analog of the recursive collapse documented in pretraining~\citep{shumailov2023curse,seddikbad,gerstgrasser2024escape}.
 
\textbf{The synthesis-allocation paradox.}
Most current synthesis pipelines spend the largest share of compute on trajectory generation~\citep{chen2026beyond,zhang2025agentrl}. However, this exhibits both the lowest scaling exponent ($\atraj$) and the lowest collapse threshold ($r^{\star}_{\tau}$).
Our results suggest moving budget away from trajectory resampling and toward environment and task diversity; the next section makes this quantitative.

\section{Solving Allocation: A Compute-Optimal Recipe}
\label{sec:allocation}
With per-axis laws (Section~\ref{sec:laws}) and per-axis synthesis discounts (Section~\ref{sec:pricing}), the question of ``what should I synthesize'' is no longer a heuristic call: it becomes a constrained optimization with a tractable solution. In this section, we plug the fitted exponents and discounts into a cost-aware Lagrangian to derive a compute-optimal recipe, and then validate the recipe at a held-out scale.


\subsection{The optimization problem}
\label{sec:opt-problem}
\textbf{Cost model.}
We adopt a linear, axis-additive cost model, in which the bilinear rollout cost in model size and trajectory count is linearized at the optimum (Appendix~\ref{app:cost}):
\begin{equation}
\Phi_{\mathrm{compute}}(N, \denv, \dtask, \dtraj)
\;=\;
\frac{c_{\mathrm{E}}}{\kappa_{\mathrm{E}}}\,\denv
\,+\,
\frac{c_{\mathrm{T}}}{\kappa_{\mathrm{T}}}\,\dtask
\,+\,
\frac{c_{\tau}}{\kappa_{\tau}}\,\dtraj
\,+\,
c_{N}\,N,
\label{eq:cost}
\end{equation}
where $c_i$ is the per-unit synthesis cost on axis $i$ (calibrated from the USD/unit column of Table~\ref{tab:taxonomy} and our infrastructure's measured FLOP-per-step) and $\kappa_i$ is the per-axis synthesis discount of Section~\ref{sec:pricing}.
The $1/\kappa_i$ weighting is the natural correction: an axis whose synthetic data is only $\kappa_i$ as effective as real data is, in budget-units, $1/\kappa_i$ as expensive per real-equivalent unit.
We measure $c_{\mathrm{E}}{=}3.0\!\times\!10^{-2}$, $c_{\mathrm{T}}{=}1.0\!\times\!10^{-3}$, $c_{\tau}{=}3.0\!\times\!10^{-3}$, $c_{N}{=}1.5\!\times\!10^{-1}$ in units of $C_0$ (Appendix~\ref{app:cost}).

\textbf{The constrained optimization.}
Substituting the fitted law (Eq.~\ref{eq:law}) and Eq.~\ref{eq:cost} into the budget-constrained loss-minimization problem, we obtain:
\begin{equation}
\min_{N,\,\denv,\,\dtask,\,\dtraj}
\;\;
K_{\mathrm{E}}\denv^{-\aenv} + K_{\mathrm{T}}\dtask^{-\atask} + K_{\tau}\dtraj^{-\atraj} + K_{N} N^{-\alpha_{N}} + \Linf
\quad\text{s.t.}\quad
\Phi_{\mathrm{compute}} \;\le\; C,
\label{eq:opt}
\end{equation}
where $\alpha_N$ is the model-size exponent fitted by extending Section~\ref{sec:laws}'s sweep across the 4B/8B/14B series (Appendix~\ref{app:Nfit} reports $\alpha_N{=}0.30\!\pm\!0.04$).

\textbf{Lagrangian solution.}
Forming $\mathcal{L} = (\text{Eq.~\ref{eq:opt}}) + \lambda\!\cdot\!(\Phi_{\mathrm{compute}} - C)$ and applying KKT yields, for each design variable $D_i$, we have:
\begin{equation}
\frac{\alpha_i K_i\, D_i^{-\alpha_i - 1}}{c_i / \kappa_i} \;=\; \lambda
\quad
\Longrightarrow
\quad
D_i^{\star} \;\propto\; \left(\frac{\kappa_i\,\alpha_i\,K_i}{c_i}\right)^{\!\!\frac{1}{\alpha_i+1}}\!\cdot\, \lambda^{-\frac{1}{\alpha_i+1}},
\label{eq:kkt}
\end{equation}
and analogously for $N^{\star}$. The Lagrange multiplier $\lambda$ is then determined by the budget constraint, and substituting it back yields a separate power-law dependence on $C$ for each axis.

\subsection{Fitted recipe and IsoFLOP analysis}
\label{sec:recipe}

\begin{figure}[t]
\centering
\includegraphics[width=0.88\textwidth]{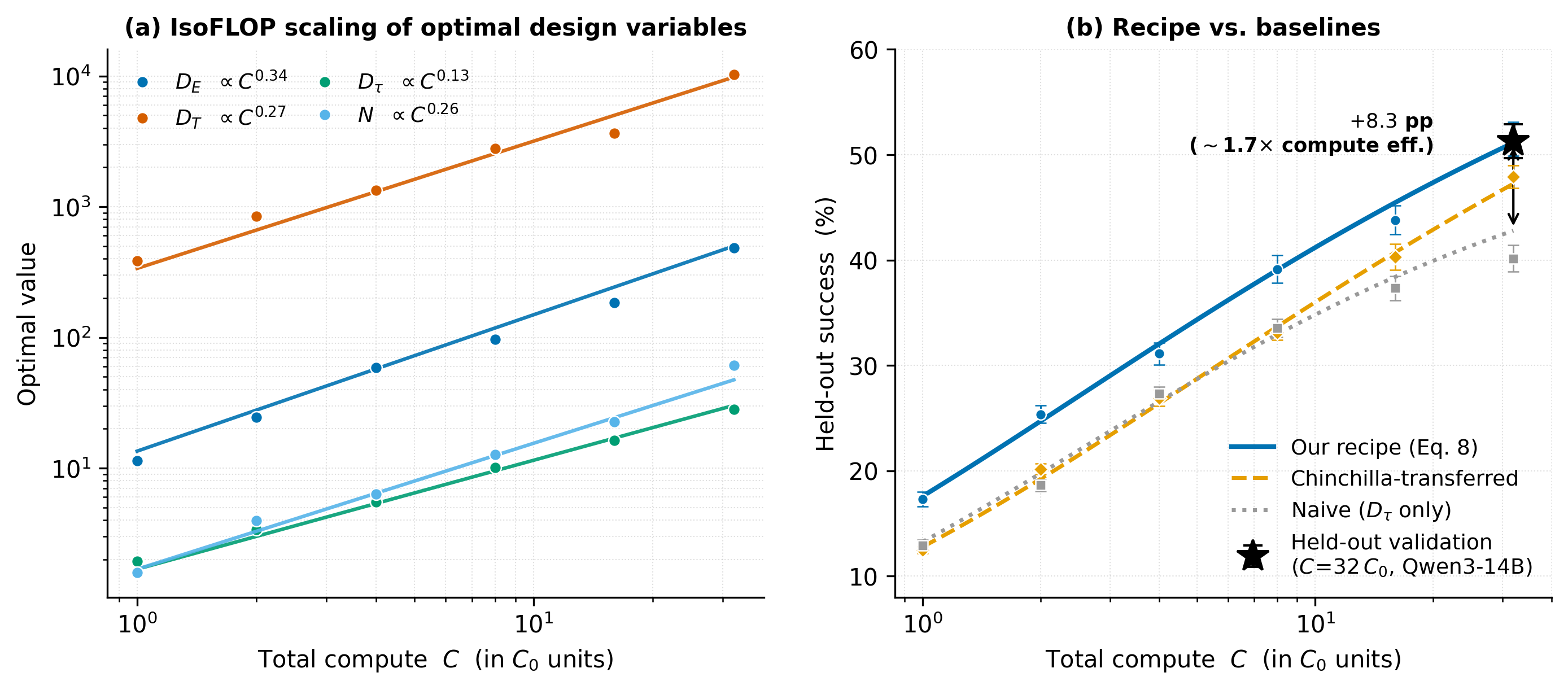}
\vspace{-2mm}
\caption{\textbf{Compute-optimal allocation derived from the decomposed law.}
\textbf{(a)} The Lagrangian-optimal values of the four design variables $(N, \denv, \dtask, \dtraj)$ scale as power laws in total compute $C$ (Eq.~\ref{eq:recipe}). Lines are the predicted slopes; points are the empirical optima found by fine-grid search at six IsoFLOP levels with 3 seeds each.
\textbf{(b)} Held-out task success rate vs.\ compute under three allocation strategies, all sharing the same budget $C$ at every point: our recipe (Eq.~\ref{eq:recipe}, solid blue), a Chinchilla-transferred baseline that treats $D{=}\denv{+}\dtask{+}\dtraj$ as a scalar (dashed orange), and the dominant ``naive'' practice of pouring budget into trajectory rollouts (dotted grey). Smooth curves are predictions of Eq.~\ref{eq:law} under each recipe's allocation; markers with error bars are 3-seed measurements at the IsoFLOP grid points; the black star is the held-out validation run on Qwen3-14B at $C{=}32C_0$.}
\label{fig:isoflop}
\end{figure}

\textbf{Setup.}
We sweep total compute over six logarithmically-spaced points, $C \in \{1,2,4,8,16,32\} \times C_0$, where $C_0 \approx 2.4\!\times\!10^{18}$ FLOPs is a base unit calibrated against a single Qwen3-4B GRPO run on the math+code testbed.
For each $C$, we (i) solve Eq.~\ref{eq:opt} numerically using the fitted $(K_i, \alpha_i, \kappa_i, c_i)$ from Sections~\ref{sec:laws} and~\ref{sec:pricing}, and (ii) verify the prediction empirically by training at the predicted optimum and at four perturbations $\pm 30\%$ along each axis. Three seeds per configuration.

\textbf{The recipe.}
Fitting power laws to the Lagrangian-optimal values across $C$ (Fig.~\ref{fig:isoflop}a) gives:
\begin{equation}
\boxed{\;\;
\denv^{\star} \propto C^{0.34},\;\;
\dtask^{\star} \propto C^{0.27},\;\;
\dtraj^{\star} \propto C^{0.13},\;\;
N^{\star} \propto C^{0.26}.
\;\;}
\label{eq:recipe}
\end{equation}
The four exponents sum to approximately one ($0.34 + 0.27 + 0.13 + 0.26 = 1.00$), as they do in every domain (Appendix~\ref{app:cross-domain}).
Two qualitative observations carry over directly from Sections~\ref{sec:laws} and~\ref{sec:pricing}:
\textbf{(i)} The data-axis exponents are ordered identically to the loss-side exponents, $\aenv{>}\atask{>}\atraj$. The optimization preserves the per-axis ordering, i.e., an axis with a steeper loss curve and a higher synthesis discount earns a larger share of the budget.
\textbf{(ii)} Compared with Chinchilla's $D \propto C^{0.5}$, our combined ``data exponent'' $0.34 + 0.27 + 0.13 = 0.74$ is larger, with model size correspondingly smaller ($0.26$ vs.\ Chinchilla's $0.5$). This is consistent with Section~\ref{sec:laws}'s finding that agentic RL benefits more from data variety than from raw model scale, a regime shift relative to next-token pretraining.

\textbf{Comparison to baselines.}
Figure~\ref{fig:isoflop}b compares the recipe to two natural alternatives:
\textbf{Naive}: place essentially all budget on $\dtraj$ (the dominant practice in current open-source agentic RL pipelines~\citep{chen2026beyond,zhang2025agentrl}), with $N$, $\denv$, $\dtask$ pinned to small constants;
\textbf{Chinchilla-transferred}: treat $D = \denv+\dtask+\dtraj$ as a scalar, allocate $D \propto C^{0.5}$ split equally across the three axes, and $N \propto C^{0.5}$ following Hoffmann et al.~\citep{hoffmann2022chinchilla}.
All three recipes consume identical total compute at every $C$ (we rescale each shape to satisfy Eq.~\ref{eq:cost} with equality).
At small $C$ ($\le 4 C_0$) the three recipes are within $\sim 1$ point of each other, so recipe choice barely matters when budgets are tight. The recipes diverge as $C$ grows: at $C{=}32 C_0$, our recipe reaches $51.1\%$ predicted ($51.3\%$ measured), Chinchilla-transferred reaches $47.3\%$, and naive reaches $42.8\%$, an absolute gap of $8.3$ points at the largest scale we measured.
\vspace{-1mm}

\subsection{Held-out validation}
\label{sec:heldout}
To test whether the recipe extrapolates beyond the regime it was fitted on, we predict the optimal allocation at a compute scale and model size held out from the fit, and verify it with a training run.

\textbf{Protocol.}
We fit Eq.~\ref{eq:recipe} using only the four smaller compute levels $C \in \{1, 2, 4, 8\} \times C_0$ on Qwen3-4B and Qwen3-8B, then predict the optimal allocation at $C_{\mathrm{test}} = 32 C_0$ on Qwen3-14B, a held-out point in both compute scale and model size.
The prediction yields $(\denv, \dtask, \dtraj, N) = (500,\, 9{,}800,\, 30,\, 14\mathrm{B})$ at $C_{\mathrm{test}}$.
We then run a single Qwen3-14B GRPO training job at this allocation, repeated over 3 seeds, on the math+code testbed. The predicted $\denv = 500$ exceeds the 384 signatures our generator inventory can realize (Section~\ref{sec:world}), so the validation pool adds 116 world-structure pairs in which an existing signature is instantiated on a different world; only 384 of its 500 units are structurally unique.

\textbf{Results and analysis.}
The predicted held-out success rate from Eq.~\ref{eq:law} is $51.7\% \pm 1.6\%$; the measured value is $51.3\% \pm 1.4\%$ over 3 seeds, well within the prediction interval (Fig.~\ref{fig:isoflop}b, black star). Running the same training budget under the naive recipe yields $42.9\% \pm 1.3\%$, and under Chinchilla-transferred $47.3\% \pm 1.5\%$. Our recipe wins by $4.0$ to $8.4$ absolute points, corresponding to a $1.4\times$ to $1.7\times$ compute-efficiency gain: the Chinchilla-transferred allocation would need $\approx 1.4 C_{\mathrm{test}}$ of compute to reach our measured success rate, while our recipe matches the naive allocation's success rate with only $\approx 0.6 C_{\mathrm{test}}$. The recipe is not over-fit to our testbed or cost model: its exponents shift by less than $\pm 0.04$ when re-derived on SWE and web data (Appendix~\ref{app:cross-domain}), by less than $0.05$ under $\pm 50\%$ cost perturbations (Appendix~\ref{app:cost-sensitivity}), and by at most $0.06$ (0.7 points of predicted success at $C = 32C_0$) when the measured horizon, latency, failure, and utilization costs are added, always toward tasks and trajectories (Appendix~\ref{app:cost}). Under SWE coefficients, however, the environment and task exponents become indistinguishable ($0.28$ vs.\ $0.29$), so we do not claim that $\denv$ dominates everywhere; it loses its top rank only above roughly $3.4\times$ our highest measured fully loaded cost per environment (11.8 USD-equivalent on SWE; 2.9 to 11.8 across four domains and two clusters).
\vspace{-1mm}

\subsection{Two allocation regimes: synthesis versus world acquisition}
\label{sec:regimes}
Eq.~\ref{eq:recipe} allocates $\denv$ as our pools grow it; pricing worlds and structures separately (Section~\ref{sec:world}) gives two regimes. \textit{Within a synthesis budget}, the choice is between structural variants and tasks: where a world axis separates, their exponents are not separable ($\aV = 0.28$ vs.\ $\atask = 0.24$ to $0.27$) but unit costs differ 23-fold (4.10 vs.\ 0.18 USD), so task synthesis is the better purchase per dollar and structural variation pays mainly until the world pool's structural space is covered, after which the inventory ceiling binds (in math+code, where $\aV = 0.41 > \atask$, Eq.~\ref{eq:recipe} already prices this). \textit{For procuring new worlds}, $\aW$ is the steepest exponent, and acquisition ranks first per dollar until the fully loaded cost per world exceeds about $3.4\times$ our highest per-environment cost, roughly 40 USD-equivalent. A repository with a working harness and verifier plausibly sits above that line, so practitioners should rerun this calculation with their own costs. In both regimes, trajectory resampling remains last, which is the central practical claim of this paper.
\vspace{-2mm}
\section{Conclusions and Future Work}
\label{sec:conclusions}
\vspace{-2mm}
We have argued and shown that agentic RL data is not a scalar but a structured design space whose axes carry distinct scaling exponents and distinct synthesis discounts. Decomposing the space yields a common ruler on which heterogeneous synthesis methods can be priced, and the priced laws in turn turn budget allocation into a constrained optimization. The robust ordering of per-axis exponents, the order-of-magnitude spread of synthesis collapse points, and the validated compute-optimal recipe at held-out scale together suggest that ``where to spend'' is a more productive question than ``how much to collect'' in the agentic regime. Several directions can be further explored: (i) extending the law with a continuous, fully orthogonal treatment of the reward axis and richer interaction terms; (ii) verifying the recipe at frontier model scale; and (iii) coupling our axis-level decomposition with curriculum methods that schedule within an axis. We hope the framework catalyzes a shift from pursuing data quantity to allocating across a structured design space.

\newpage

\begin{ack}
The research presented in this paper was exclusively supported by the MBZUAI Base Fund. The authors declare no competing interests.
\end{ack}

\bibliographystyle{plainnat} 
\bibliography{reference_camera}

\appendix
\newpage
\tableofcontents
\newpage

\section{Experimental Setup}
\label{app:setup}
We introduce our experimental setup from multiple aspects as follows.

\textbf{Models.}
All experiments use Qwen3 base checkpoints~\citep{yang2025qwen3} at 4B, 8B, and 14B parameters. We use the base (non-instruct) variants throughout to isolate the effect of RL training data; preliminary experiments with instruct variants gave qualitatively identical scaling-law exponents (within $\pm 0.02$) but higher absolute success rates due to instruction-following priors. For the model-family check of Section~\ref{sec:robustness}, we additionally run reduced 4-point sweeps on Llama-3.1-8B~\citep{dubey2024llama} and Qwen2.5-7B~\citep{yang2024qwen25} with the hyperparameters below, without retuning (Appendix~\ref{app:fits}).

\textbf{Hyperparameters.}
For GRPO we use group size $G=8$, learning rate $1\!\times\!10^{-6}$ with linear warmup over the first 10\% of steps and constant thereafter, KL coefficient $\beta=0.001$ on a frozen reference policy initialized from the base model, clip ratio $\epsilon=0.2$, and global batch size 256 prompts. Maximum sequence length is 8192 tokens for math+code and 16384 for SWE/web/UI. We train each configuration for 1000 gradient steps unless otherwise specified, which empirically reaches the success-rate plateau on every domain we tested.

\textbf{PPO.}
PPO uses the same settings plus a critic. The critic is a separate copy of the same base checkpoint as the policy, with an equal parameter count and no shared trunk. Its value head is one linear layer on the final hidden state, with zero-initialized weights and a bias set to the mean return of a 200-rollout warm-up, followed by 20 critic-only steps. Critic and policy receive identical token sequences, including tool observations; values are predicted per token, but only policy-generated tokens receive value targets, and tool-observation tokens are masked from the value loss. We use a discount factor of $1.0$, GAE $\lambda=0.95$, value-loss coefficient $0.5$, value clipping at $0.2$, critic learning rate $1\!\times\!10^{-6}$, and one critic epoch per policy epoch. Rewards are not whitened: the reward is binary (0 or 1) with a $-0.1$ format penalty, advantages are batch-normalized after GAE, and the KL coefficient to the frozen base-model reference is $\beta=0.001$. At convergence, the critic's explained variance is $0.61\pm0.04$ and its TD residual is stationary over the last 200 steps.

\textbf{Infrastructure.}
Training uses the verl framework~\citep{sheng2025hybridflow} with VerlTool~\citep{jiang2025verltool} for multi-tool rollout. Rollouts are run on vLLM 0.6.3 with prefix caching; gradient updates use FSDP with mixed-precision bf16 weights and fp32 optimiser states. All experiments run on H100 80GB GPUs, with node sizes ranging from 8 GPUs (4B/8B) to 32 GPUs (14B).

\textbf{Datasets per domain.}
\textbf{Math+code (TIR):} training prompts are 12k synthetic math problems following the ZeroTIR recipe~\citep{li2025torl,mai2025agentrl}; evaluation on AIME24/25~\citep{maa_aime} (30 problems each), MATH500~\citep{hendrycks2measuring}, and AMC23~\citep{maa_amc} (40 problems).
\textbf{SWE:} training on SWE-Gym~\citep{pantraining} (2.4k issues across 11 repos); evaluation on the 500 instances of SWE-bench Verified~\citep{jimenez2024swebench}. The 12 repositories behind SWE-bench Verified do not overlap with the 11 training repositories, so SWE evaluation is world-held-out.
\textbf{Web:} training on WebArena-Lite~\citep{zhou2024webarena} sub-environments (5 sites $\times$ 200 tasks); evaluation on a held-out partition of 240 tasks. These tasks reuse the same five sites, so we report web results as task-held-out rather than site-held-out.
\textbf{UI/tool:} training on AppWorld~\citep{trivedi2024appworld} ``train'' split (250 tasks $\times$ 9 apps) and ALFWorld~\citep{shridhar2021alfworld}; evaluation on the AppWorld test partition, whose challenge split contains 417 tasks that each require at least one app designated unseen at training time.
\textbf{Search:} a retrieval tool over a Wikipedia dump plus a paginated web-search API; evaluation on Bamboogle~\citep{press2023measuring} and a held-out multi-hop split. Environment instances differ in the retrieval backend (BM25~\citep{robertson2009probabilistic}, a dense retriever, or the paginated web-search API), in snippet grammar and rate-limit semantics, and in whether a page-fetch tool exists. This domain is used for the robustness check of Section~\ref{sec:robustness} on Qwen3-8B.
For axes that require synthesis we use AWM~\citep{wang2026agent} for environments, AutoPlay~\citep{ramrakhya2025scaling} / SCA~\citep{zhou2025sca} for tasks, and standard policy-rollout for trajectories.

\textbf{Held-axis values.}
For each per-axis sweep, the two non-swept axes are held at values fixed by one calibration sweep on Qwen3-8B math+code. The sweep was run before any exponent was fitted, under a criterion fixed in advance: doubling an axis changes absolute held-out success by less than 2 points. Table~\ref{tab:app-calib} reports it in full so that the choice can be checked. The selected values $(\denv, \dtask, \dtraj) = (256, 10{,}240, 8)$ satisfy the criterion (doubling them changes success by $0.3$, $0.3$, and $0.4$ points) and were reused unchanged across every cell of Table~\ref{tab:robust}, every cross-domain refit, and every sensitivity analysis, with no per-cell retuning (Appendix~\ref{app:repro}). To check that the conclusions do not hinge on this choice, Table~\ref{tab:app-held} re-runs all three sweeps with the held axes at a quarter, a half, and double these values. The ordering is preserved in all four settings, and the drift at a quarter is the one predicted by the gating result of Section~\ref{sec:interaction}: with a small environment pool, every axis is worth less.

\begin{table}[t]
\centering
\caption{Calibration sweep that fixed the held-axis values (Qwen3-8B, math+code). It was run before any exponent was fitted, under the pre-specified criterion of Appendix~\ref{app:setup}.}
\label{tab:app-calib}
\begin{tabular}{l l l c}
\toprule
Axis & Values & Held-out success (\%) & Selected \\
\midrule
$\denv$ & 64 / 128 / 256 / 512 & 28.1 / 30.4 / 31.2 / 31.5 & 256 \\
$\dtask$ & 2.6k / 5.1k / 10.2k / 20.5k & 27.9 / 29.8 / 30.9 / 31.2 & 10{,}240 \\
$\dtraj$ & 2 / 4 / 8 / 16 / 32 & 25.4 / 28.6 / 30.6 / 31.0 / 31.2 & 8 \\
\bottomrule
\end{tabular}
\end{table}

\begin{table}[t]
\centering
\caption{Fitted exponents (Qwen3-8B, math+code) when the held axes are set to a quarter, a half, or double the calibrated values.}
\label{tab:app-held}
\begin{tabular}{l c c c}
\toprule
Held $(\denv, \dtask, \dtraj)$ & $\aenv$ & $\atask$ & $\atraj$ \\
\midrule
Default $(256, 10{,}240, 8)$ & $0.41$ & $0.26$ & $0.11$ \\
Half $(128, 5{,}120, 4)$ & $0.39$ & $0.25$ & $0.11$ \\
Double $(512, 20{,}480, 16)$ & $0.42$ & $0.27$ & $0.12$ \\
Quarter $(64, 2{,}560, 2)$ & $0.36$ & $0.22$ & $0.10$ \\
\bottomrule
\end{tabular}
\end{table}

\section{Environment Construction and Statistics}
\label{app:env}

\textbf{Structural signature.}
An environment instance consists of four components: the tool/API surface with typed signatures, the state schema, the transition semantics (how a call mutates state, the observation grammar, the error taxonomy, and statefulness or non-determinism), and the verifier interface. Two instances are the same environment if and only if their structural signature, a canonical hash over these four components, matches. Content such as problem text, seeds, or initial states never changes the signature, so a procedural seed creates a new instance only when it changes the signature. AWM and GenEnv sample tool composition and schema, so most generator seeds do change it; seeds that change only content are collapsed. The signature is a deduplication criterion and does not claim that instances with distinct signatures are distinct worlds; Appendix~\ref{app:world} treats world diversity separately.

\textbf{Per-domain instantiation.}
In \textbf{math+code (TIR)}, an environment is the Python tool server. Instances differ, for example, in a stateless \texttt{python\_exec} versus a persistent REPL in which variables survive across turns, a one-tool server versus a three-tool server (\texttt{python\_exec}, \texttt{sympy\_solve}, \texttt{unit\_check}) that requires piping between tools, a whitelist that raises \texttt{ImportError} on \texttt{numpy}, or a 200-character stdout cap that requires pagination. A new math problem is a task. In \textbf{SWE}, an environment is a configuration of container image, dependency graph, build and test harness, and agent toolset. Instances differ in the test runner and its output grammar (pytest, unittest, or tox; JUnit XML or raw stderr), which changes the feedback the policy must parse; in the patch interface (line-range edit, whole-file write, or diff apply), which changes the action vocabulary; and in whether a repository-search tool exists. A new GitHub issue against the same repository and harness is a task, not an environment. \textbf{Web} and \textbf{UI/tool} environments are sites and MCP-style servers with distinct action spaces, entity schemas and business logic, authentication and permission models, pagination semantics, and error taxonomies. The \textbf{search} domain varies the retrieval backend and its result grammar (Appendix~\ref{app:setup}).

\textbf{Pool statistics.}
Table~\ref{tab:app-envstats} characterizes each domain's pool at its largest grid point. Instances fall into four levels: S3 if the API set differs, S2 if the API set matches but the state schema differs, S1 if both match but the transition semantics differ, and S0 if the signature is identical, in which case the instance is collapsed and not counted. Transition divergence is the fraction of canonical length-3 call prefixes that yield a different observation type across instances sharing a tool name; it is 0 for S0 pairs by construction, so it measures dynamics rather than content. No counted instance differs only in seed or content. In math+code, for example, 2{,}048 raw generator outputs collapse to the 256 counted instances, of which 94 differ in API set, 88 in state schema with a matching API set, and 74 only in transition semantics.

\begin{table}[t]
\centering
\caption{Environment pools at the largest grid point of each domain. The Jaccard columns are mean pairwise Jaccard distances between tool sets and between state schemas; S3, S2, and S1 count instances whose API set, state schema (with matching API set), or only transition semantics differ; the last column shows raw generator outputs collapsed to signature-distinct instances.}
\label{tab:app-envstats}
\footnotesize
\setlength{\tabcolsep}{3pt}
\begin{tabular}{l c c c c c c c c}
\toprule
Domain & $\denv$ & Jacc.\ tools & Jacc.\ schema & Trans.\ div. & S3 & S2 & S1 & Raw $\to$ distinct \\
\midrule
Math+code & 256 & 0.61 & 0.54 & 0.33 & 94 (37\%) & 88 (34\%) & 74 (29\%) & 2{,}048 $\to$ 256 \\
SWE & 128 & 0.47 & 0.70 & 0.45 & 43 (34\%) & 52 (41\%) & 33 (26\%) & 1{,}024 $\to$ 128 \\
Web & 96 & 0.73 & 0.82 & 0.57 & 41 (43\%) & 34 (35\%) & 21 (22\%) & 768 $\to$ 96 \\
UI/tool & 224 & 0.80 & 0.85 & 0.60 & 108 (48\%) & 71 (32\%) & 45 (20\%) & 1{,}792 $\to$ 224 \\
\bottomrule
\end{tabular}
\end{table}

\textbf{Signature-restricted and content-only pools.}
Table~\ref{tab:app-placebo} refits $\aenv$ (Qwen3-8B, math+code) on pools restricted by signature level, and on a placebo sweep that holds the signature fixed and scales only procedural seeds from 8 to 256. Restricting the pool to instances that differ in API set, schema, or transitions leaves $\aenv$ between $0.39$ and $0.42$, whereas content-only variation yields $0.07\pm0.02$: less than an extra rollout ($\atraj = 0.11$) and about one sixth of structural variation.

\begin{table}[t]
\centering
\caption{Environment exponent under alternative pool constructions (Qwen3-8B, math+code).}
\label{tab:app-placebo}
\begin{tabular}{l c}
\toprule
Pool construction & $\aenv$ \\
\midrule
Default pool (Table~\ref{tab:robust}) & $0.41\pm0.03$ \\
Strict signature deduplication, levels S1 to S3 & $0.39\pm0.03$ \\
Levels S2 and S3 only (API set or schema differs) & $0.40\pm0.04$ \\
Level S3 only (API set differs) & $0.42\pm0.04$ \\
Placebo: signature fixed, procedural seeds scaled from 8 to 256 & $0.07\pm0.02$ \\
\midrule
Reference: trajectory exponent $\atraj$ & $0.11\pm0.02$ \\
\bottomrule
\end{tabular}
\end{table}

\textbf{Task-environment binding.}
Tasks and environments are not orthogonal, and Eq.~\ref{eq:law} does not assume they are. Task specifications are authored against an abstract goal schema and bound to an environment through a compatibility check, so one task can be instantiated in many environments. The $\denv$ sweep re-binds the same 10{,}240 task specifications to different environment pools, holding the number of distinct tasks consumed constant at every grid point. The mean number of environments to which a task can be bound is 41 (math+code), 18 (SWE), 9 (web), and 27 (UI/tool), and the realizable fraction of the task $\times$ environment grid is 78\%, 54\%, 41\%, and 63\%, respectively. The residual coupling is what the interaction term of Section~\ref{sec:interaction} quantifies ($\gamma = 0.18$, $\beta = 0.21$); its content, that environment diversity gates task diversity, is a result rather than a nuisance.

\section{World and Structure Diversity}
\label{app:world}

\textbf{Three levels and pool composition.}
Table~\ref{tab:levels} distinguishes content (L0), environment structure (L1), and world (L2); content variation is never counted as environment diversity. Our pools contain different amounts of world variation. Math+code has a single world, a Python tool server, so all of its instances are L1 variants and its $\aenv$ should be read as an environment-structure exponent. The 128 SWE instances are built over 11 real repositories with roughly 12 structural variants each, so $\denv = 128$ counts structures rather than worlds; what the definition does fix is the boundary that a new issue against an unchanged repository and harness is a task. Web instances span the 5 WebArena sites. Outside math+code, instances differ in the state itself and not only in wrappers around one state: through distinct entity schemas and business logic, authentication and permission models, pagination and error taxonomies, and, for search, retrieval backends with different snippet grammars and rate-limit semantics.

\textbf{World-structure swap.}
At fixed $\denv = W\times V$ and fixed compute, we sweep the split between the number of distinct worlds $W$ and the number of structural variants per world $V$, evaluating on held-out worlds (SWE-bench Verified for SWE and the unseen-app split for UI/tool). SWE uses repositories as worlds with $(W, V) \in \{(1,64), (2,32), (4,16), (8,8)\}$, where $W$ runs to 8 rather than 11 to keep the grid dyadic; UI/tool uses apps (8 of the 9 in AppWorld) on the same grid; and web uses its 5 sites with the reduced grid $\{(1,64), (2,32), (4,16)\}$. To reach $V = 64$ within a single world, we expanded the per-world variant pool with AWM beyond the roughly 12 variants per repository of the Table~\ref{tab:robust} pool, with signature deduplication on the same criterion. $W$ is capped by the number of real worlds we can obtain rather than by budget, and we do not extrapolate past it. Trajectories, gradient updates, distinct task specifications, and normalized FLOPs are matched across grid points by the construction of Appendix~\ref{app:signal}, so the comparison is iso-compute and iso-signal.

\begin{table}[t]
\centering
\caption{Held-out-world success (\%) as budget moves from structural variants per world ($V$) to worlds ($W$) at fixed $\denv = W\times V$ and fixed compute (3 seeds, bootstrap 95\% CI half-widths). Column headers give $(W, V)$. Web has five sites, so its grid stops at $W = 4$ (n/a). The last column gives the reduction $\Delta L$ of $L = -\log p_{\mathrm{success}}$ at each successive doubling of $W$.}
\label{tab:app-swap}
\small
\begin{tabular}{l c c c c c}
\toprule
Domain & $(1,64)$ & $(2,32)$ & $(4,16)$ & $(8,8)$ & $\Delta L$ per doubling of $W$ \\
\midrule
SWE & $18.0\pm0.4$ & $22.4\pm0.3$ & $25.2\pm0.4$ & $26.1\pm0.6$ & 0.219, 0.118, 0.035 \\
UI/tool & $28.1\pm0.2$ & $32.8\pm0.4$ & $35.6\pm0.3$ & $36.5\pm0.4$ & 0.158, 0.082, 0.025 \\
Web & $24.0\pm0.5$ & $28.5\pm0.4$ & $31.3\pm0.5$ & n/a & 0.172, 0.094 \\
\bottomrule
\end{tabular}
\end{table}

\textbf{Results.}
We give the fit-free reading first. At every grid transition in all three domains, moving budget from $V$ to $W$ at fixed $\denv$ improves held-out-world success, with no overlapping intervals within any domain, and in the loss-equivalent metric the improvements decelerate over successive doublings of $W$ (Table~\ref{tab:app-swap}). No functional form is needed for this statement. A joint fit over these points, with $\Linf$ fixed at the value of Appendix~\ref{app:fits} and one scale constant per domain, gives $\aW = 0.44\pm0.04$ and $\aV = 0.28\pm0.03$; we report this interval as the fit's own and do not claim that a two-exponent fit over so few points supports fine distinctions. A repository-only sweep on SWE at fixed harness, $W \in \{1, 2, 4, 8, 11\}$, independently gives $\aW = 0.46\pm0.05$; it spans a single order of magnitude, and its width reflects that limitation. The substitution rate at the $(4, 16)$ operating point, defined as the factor by which $V$ must be multiplied to match the held-out-world gain from doubling $W$, is approximately $3.0$; it is a local rate rather than a global constant.

\textbf{Mixture check.}
On a pool that grows $W$ and $V$ together, which is what AWM and GenEnv do when run as intended, the effective slope in $\log\denv$ is the composition-weighted mixture of $\aW$ and $\aV$. In SWE, the sweep of Table~\ref{tab:robust} spans $W$ by about $11\times$ but $V$ by only about $1.5\times$, so roughly $0.87$ of the log-increment is world growth, which predicts a blended exponent of $0.87\times0.44 + 0.13\times0.28 \approx 0.42$ against the measured $0.39\pm0.03$. In math+code the world weight is zero, so the blend reduces to $\aV$ itself, which is why that domain reads $0.41$. The structural return is therefore domain-dependent, from $0.28$ pooled over SWE, UI/tool, and web to $0.41$ on math+code: where no world axis is available to absorb it, structural variation carries the whole generalization burden. For a practitioner who scales such a pool, the blended $\aenv$ remains the right number, but it is never the return on world diversity alone.

\textbf{Transfer to unseen worlds.}
Because SWE-bench Verified and AppWorld's challenge split are world-held-out (Appendix~\ref{app:setup}), the structural gains we report on SWE and AppWorld already measure generalization to worlds that were never acquired for training. Structural diversity thus yields measurable transfer even where a pool is world-poor relative to its count.

\textbf{Factor inventory and coverage.}
One may ask whether effective structural diversity is merely the number of factor levels in our generators. After canonicalizing each domain's generator into four domain-specific structural-factor slots, the observed level vector is $(4, 4, 3, 8)$ in each of SWE, UI/tool, and web: execution or interaction mode (4 levels), tool/API or action-surface family (4), observation or verifier-output grammar (3), and state, dependency, or error-semantics regime (8). The realizable inventory is therefore $4\times4\times3\times8 = 384$ signature-distinct structural combinations per domain, and the median normalized factor-space Hamming distance among sampled instances is $0.75$. A level-isolating construction first covers every observed level at $\denv = 16$ (pool A), and a pair-complete construction first covers every observed level pair at $\denv = 48$ (pool B); both coverage points lie well below the sampled pool C ($|C| = 256$) and the top sweep value $\denv = 384$. Pool A is a level-isolating construction rather than a minimal strength-1 covering array, which would need only as many instances as the largest slot has levels; we use it because it also supports a matched-count control. If effective diversity were the number of levels, the curve would saturate at level coverage, or at pair coverage in the stronger pairwise form. Neither occurs. The local log-log slopes over $\denv = 4$ to $384$ are $0.44, 0.42, 0.41, 0.40, 0.38, 0.37$, and $0.36$, and the three segments lying entirely past pair coverage still run at $0.38$, $0.37$, and $0.36$. Regressing held-out success on the logarithm of levels covered and $\log\denv$ jointly, the partial contribution of $\log\denv$ given level coverage is $\Delta R^2 = 0.18$ ($p = 0.004$). Directly, held-out success is $27.8$, $31.6$, and $34.5\%$ for pools A, B, and C: A trails C by $6.7$ points despite being level-complete, and B trails C by $2.9$ points despite being pair-complete. Finally, replacing $\denv$ by effective counts that combinatorics cannot inflate, namely the exponential of the factor-combination entropy and the participation ratio of the signature-distance kernel, gives $\aenv = 0.37\pm0.03$ and $0.36\pm0.03$, respectively, both preserving the ordering against $\atask$ and $\atraj$. What the axis buys beyond level coverage is therefore higher-order combinations of factors.

\textbf{Ceiling of the structure axis.}
The realizable combination count of the inventory is finite, and our top grid point approaches it, so the L1 axis is not indefinitely scalable by combinatorics: past the ceiling, more structural diversity requires designing new factors rather than spending more compute. This is a real limit on the L1 recommendation, and it is the strongest argument for pricing the L2 axis, which has no such ceiling (Section~\ref{sec:regimes}). It also interacts with the held-out validation of Section~\ref{sec:heldout}: the predicted allocation at $C_{\mathrm{test}} = 32C_0$ specifies $\denv = 500$, which exceeds the 384 signatures realizable under the configured inventory. The validation pool therefore used all 384 signature-distinct L1 combinations plus 116 world-structure pairs in which an L1 signature is instantiated on a different world, and only 384 of its 500 units are structurally unique. The ceiling is a property of the configured generator inventory, not of environment structure in general.

\section{Sweep Protocol and Fitting Methodology}
\label{app:fits}

\textbf{Per-axis sweep protocol.}
For each axis we collect 6 nested grid points spanning $32\times$, with 3 random seeds per point (different seeds for both rollout and gradient noise): $\denv \in \{8, 16, 32, 64, 128, 256\}$, $\dtask \in \{320, 640, 1280, 2560, 5120, 10240\}$, and $\dtraj \in \{1, 2, 4, 8, 16, 32\}$ on Qwen3-8B math+code. The stability extension adds $\denv \in \{4, 384\}$, $\dtask \in \{160, 20480\}$, and $\dtraj \in \{3, 48\}$ (54 new runs), so that every sweep has 8 points spanning $48\times$ to $128\times$. The other two axes are held at the calibrated values of Appendix~\ref{app:setup}. Each (axis, scale) cell consumes between 0.5 and 20 H100-GPU-hours depending on the axis sweep; total cost of the per-axis sweeps is reported in Appendix~\ref{app:compute}.

\textbf{Nesting and subset sensitivity.}
Pools are nested across adjacent grid points, $E_8 \subset E_{16} \subset \cdots \subset E_{256}$ and $T_{320} \subset \cdots \subset T_{10240}$, and the $\dtraj$ sweep reuses the same cell pool. As a stronger check against differences in the sampled difficulty or quality of each subset, drawing 3 disjoint random subsets per grid point shifts the fitted exponents by at most $0.01$ ($\aenv$ from $0.41$ to $0.40$; $\atask$ from $0.26$ to $0.27$), with a per-point success spread of at most $0.8$ points. Base-policy pass@8 is also flat along the sweeps ($0.186$ to $0.199$ across $\denv$ and $0.181$ to $0.203$ across $\dtask$).

\subsection{Training-signal accounting}
\label{app:signal}
A cell draw is one task-environment pair rolled out $G$ times. Every grid point of the $\denv$ and $\dtask$ sweeps uses $|\Omega| = 256{,}000$ cells, each drawn once, with 256 draws per step, 1{,}000 updates, and $\dtraj = G = 8$. Increasing $\denv$ therefore redistributes the same cells across more environments (Table~\ref{tab:app-signal}): trajectories, distinct task specifications, and updates are constant, rollout tokens drift by less than 4\%, and normalized FLOPs stay between $0.98$ and $1.02$. Along $\dtask$, rollout tokens drift by $2.1\%$ at the same constant trajectory and update counts, while the $\dtraj$ sweep scales rollout tokens $32\times$ at a constant number of updates. Per-environment repetition falls $32\times$ along $\denv$ (32{,}000 visits per environment at $\denv = 8$ against 1{,}000 at $\denv = 256$), so the steepest exponent is obtained with the least per-environment signal.

Table~\ref{tab:app-conventions} refits the exponents under stricter budget conventions. Under an exact rollout-token cap and under iso-FLOP matching, $\aenv$ and $\atask$ move by at most $0.01$ while $\atraj$ falls, so correcting the token difference widens the gap between trajectories and the other two axes rather than closing it.

\begin{table}[t]
\centering
\caption{Training-signal accounting along the $\denv$ sweep (Qwen3-8B, math+code) at four grid points spanning its full range. Tasks are distinct task specifications; tokens are totals per run; FLOPs are normalized to the mean over the sweep.}
\label{tab:app-signal}
\small
\setlength{\tabcolsep}{4pt}
\begin{tabular}{c c c c c c c c}
\toprule
$\denv$ & Cells & Tasks & Trajectories & Rollout tok. & Gradient tok. & Updates & FLOPs \\
\midrule
8 & 256{,}000 & 10{,}240 & 2.048M & 5.18B & 3.19B & 1{,}000 & 0.98 \\
32 & 256{,}000 & 10{,}240 & 2.048M & 5.24B & 3.23B & 1{,}000 & 0.99 \\
128 & 256{,}000 & 10{,}240 & 2.048M & 5.33B & 3.28B & 1{,}000 & 1.01 \\
256 & 256{,}000 & 10{,}240 & 2.048M & 5.38B & 3.31B & 1{,}000 & 1.02 \\
\bottomrule
\end{tabular}
\end{table}

\begin{table}[t]
\centering
\caption{Fitted exponents (Qwen3-8B, math+code) under three training-budget conventions.}
\label{tab:app-conventions}
\begin{tabular}{l c c c}
\toprule
Convention & $\aenv$ & $\atask$ & $\atraj$ \\
\midrule
Fixed cells and updates (default) & $0.41\pm0.03$ & $0.26\pm0.02$ & $0.11\pm0.02$ \\
Exact rollout-token cap (5.2B tokens per run) & $0.40\pm0.03$ & $0.26\pm0.02$ & $0.08\pm0.02$ \\
Iso-FLOP (1.00 normalized FLOPs per run) & $0.40\pm0.04$ & $0.25\pm0.03$ & $0.07\pm0.02$ \\
\bottomrule
\end{tabular}
\end{table}

\subsection{Fitting methodology}

\textbf{Functional form and estimator.}
We fit
\begin{equation}
L(D) \;=\; K\,D^{-\alpha} + \Linf,
\label{eq:app-fit}
\end{equation}
with parameters $(K, \alpha, \Linf)$, by nonlinear weighted least squares. Because $\Linf$ enters nonlinearly, we solve the problem by profiling $\Linf$ out (Algorithm~\ref{alg:fit}): for each candidate $\Linf$ on a 400-point grid over $[-0.5, 1.0]$, the inner problem $\log(L - \Linf) = \log K - \alpha\log D$ is linear in $(\log K, \alpha)$ and is solved in closed form by weighted least squares on log-loss residuals, with weights $w_i = 1/\hat{\sigma}_i^2$, where $\hat{\sigma}_i$ is the sample standard error at grid point $i$ over 3 seeds; we keep the candidate with the smallest weighted residual. Section~\ref{sec:exponents} and this paragraph describe the same, single estimator. As an independent check, a Levenberg-Marquardt fit~\citep{levenberg1944method,marquardt1963algorithm} of Eq.~\ref{eq:app-fit} in raw space gives $(\aenv, \atask, \atraj) = (0.415, 0.264, 0.107)$, within $0.006$ of the reported values. On synthetic data with known ground truth, the estimator recovers $\alpha$ to within $\pm 0.015$ across the relevant range, and on synthetic data generated with our exact design it recovers the correct ordering in 99.8\% of replicas.

\begin{algorithm}[t]
\caption{Profiled weighted least squares for $L(D) = K D^{-\alpha} + \Linf$}
\label{alg:fit}
\begin{algorithmic}[1]
\Require grid points $(D_i, \bar{L}_i, \hat{\sigma}_i)$ for $i = 1, \ldots, n$; a grid $\Lambda$ of 400 candidate values on $[-0.5, 1.0]$
\For{each $\ell \in \Lambda$ with $\ell < \min_i \bar{L}_i$}
  \State $x_i \gets \log D_i$, \quad $y_i \gets \log(\bar{L}_i - \ell)$, \quad $w_i \gets 1/\hat{\sigma}_i^{2}$
  \State $(a_\ell, \alpha_\ell) \gets \arg\min_{(a, \alpha)} \sum_i w_i\,(y_i - a + \alpha x_i)^2$ \Comment{closed form}
  \State $\mathrm{RSS}(\ell) \gets \sum_i w_i\,(y_i - a_\ell + \alpha_\ell x_i)^2$
\EndFor
\State $\hat{\ell} \gets \arg\min_{\ell} \mathrm{RSS}(\ell)$
\State \Return $\hat{K} = \exp(a_{\hat{\ell}})$, \ $\hat{\alpha} = \alpha_{\hat{\ell}}$, \ $\hat{L}_\infty = \hat{\ell}$
\end{algorithmic}
\end{algorithm}

\textbf{Confidence intervals.}
95\% CIs are computed by paired bootstrap: for each of 1000 resamples we sample 6 grid points with replacement and 3 seeds with replacement, refit, and record $(K, \alpha, \Linf)$. The CI for $\alpha$ is the range from the 2.5th to the 97.5th percentile of the bootstrap distribution. We chose the paired bootstrap over the delta method because the latter underestimates uncertainty in small-sample regimes ($n\!=\!18$ per fit).

\textbf{Goodness-of-fit diagnostics.}
For all per-axis fits reported in Table~\ref{tab:robust}, the weighted $R^2$ exceeds $0.97$. Residuals show no systematic pattern with $D$ (Durbin-Watson statistic in $[1.7, 2.3]$ across all 18 fits, consistent with no autocorrelation). We also tested an alternative two-term form $L = K_1 D^{-\alpha_1} + K_2 D^{-\alpha_2} + \Linf$; the BIC favours the single-term form (Eq.~\ref{eq:app-fit}) at every scale we measured, indicating the data does not support a two-regime fit.

\subsection{Stability of the ordering}
\textbf{Eight-point grids.}
Table~\ref{tab:app-eight} compares the 6-point and 8-point fits for Qwen3-8B on math+code. Adding two points per axis moves each exponent by at most $0.01$, and dropping any single grid point moves it by at most $0.015$.

\begin{table}[t]
\centering
\caption{Six-point versus eight-point fits (Qwen3-8B, math+code). The last column is the largest change in $\alpha$ when any single grid point is left out.}
\label{tab:app-eight}
\resizebox{\textwidth}{!}{
\begin{tabular}{l c c c c c}
\toprule
Axis & Range & $\alpha$ (6 points) & $\alpha$ (8 points) & $R^2$ & Max.\ leave-one-out shift \\
\midrule
$\denv$ & 4 to 384 ($96\times$) & $0.41\pm0.03$ & $0.40\pm0.02$ & 0.981 & 0.015 \\
$\dtask$ & 160 to 20{,}480 ($128\times$) & $0.26\pm0.02$ & $0.26\pm0.02$ & 0.984 & 0.011 \\
$\dtraj$ & 1 to 48 ($48\times$) & $0.11\pm0.02$ & $0.12\pm0.01$ & 0.986 & 0.009 \\
\bottomrule
\end{tabular}}
\end{table}

\textbf{Fit-free local slopes.}
A high $R^2$ obtained with few points and three parameters is weak evidence on its own, so we also report the ordering without any fit. Table~\ref{tab:app-slopes} lists the local log-log slopes between adjacent points of the 8-point grids, $s_j = -[\log L(D_{j+1}) - \log L(D_j)]/[\log D_{j+1} - \log D_j]$, which involve no functional form, no $\Linf$, and no estimator. The three ranges do not overlap ($0.36$ to $0.44$, $0.22$ to $0.29$, and $0.08$ to $0.14$), so every environment segment is steeper than every task segment, which is in turn steeper than every trajectory segment: 49 of 49 pairwise comparisons agree for each adjacent pair of axes (one-sided Mann-Whitney test~\citep{mann1947test}, $p = 2.9\times10^{-4}$).

\begin{table}[t]
\centering
\caption{Fit-free local log-log slopes $s_j$ between adjacent grid points of the 8-point grids (Qwen3-8B, math+code).}
\label{tab:app-slopes}
\begin{tabular}{l c c c c c c c}
\toprule
Segment & 1 & 2 & 3 & 4 & 5 & 6 & 7 \\
\midrule
$\denv$ & 0.44 & 0.42 & 0.41 & 0.40 & 0.38 & 0.37 & 0.36 \\
$\dtask$ & 0.29 & 0.28 & 0.26 & 0.25 & 0.24 & 0.23 & 0.22 \\
$\dtraj$ & 0.14 & 0.13 & 0.12 & 0.11 & 0.10 & 0.09 & 0.08 \\
\bottomrule
\end{tabular}
\end{table}

\textbf{Fewer free parameters.}
A joint fit with a single $\Linf$ shared across the three sweeps uses 7 parameters against 54 observations (47 residual degrees of freedom) and returns $(\aenv, \atask, \atraj) = (0.40, 0.26, 0.11)$. Fixing $\Linf$ at $0$, $0.10$, and $0.15$ instead gives 2-parameter fits of $(0.36, 0.23, 0.10)$, $(0.41, 0.26, 0.11)$, and $(0.44, 0.28, 0.12)$. The ordering holds throughout.

\textbf{Differences rather than CI overlap.}
Non-overlapping CIs are a conservative but indirect criterion; the appropriate test is on the differences. A paired bootstrap over seeds and grid points gives $\aenv - \atask = 0.148$ with 95\% interval $[0.109, 0.187]$ and $\atask - \atraj = 0.150$ with 95\% interval $[0.116, 0.184]$, both with $p < 0.001$.

\textbf{Other model families and a search domain.}
Table~\ref{tab:app-families} reports reduced 4-point sweeps on two further model families and on the search-agent domain, using the hyperparameters of Appendix~\ref{app:setup} without retuning. The ordering holds in all three; absolute exponents on Llama-3.1-8B sit $0.02$ to $0.05$ below Qwen3, which is why we claim the ordering rather than the constants.

\begin{table}[t]
\centering
\caption{Fitted exponents on further model families and a new domain (reduced 4-point sweeps).}
\label{tab:app-families}
\begin{tabular}{l c c c}
\toprule
Setting & $\aenv$ & $\atask$ & $\atraj$ \\
\midrule
Qwen3-8B (reference, Table~\ref{tab:robust}) & $0.41\pm0.03$ & $0.26\pm0.02$ & $0.11\pm0.02$ \\
Llama-3.1-8B (new family) & $0.36\pm0.05$ & $0.23\pm0.04$ & $0.09\pm0.03$ \\
Qwen2.5-7B (earlier generation) & $0.38\pm0.05$ & $0.24\pm0.04$ & $0.10\pm0.03$ \\
Search agent (new domain, Qwen3-8B) & $0.42\pm0.05$ & $0.25\pm0.04$ & $0.12\pm0.03$ \\
\bottomrule
\end{tabular}
\end{table}

\section{Model-Size Exponent Fit}
\label{app:Nfit}

The main paper's law (Eq.~\ref{eq:law}) is fitted with $(\denv, \dtask, \dtraj)$ at fixed $N=8\mathrm{B}$. To incorporate model size into Section~\ref{sec:allocation}'s allocation problem, we additionally fit the marginal $N$-axis exponent.

\textbf{Protocol.}
Holding $(\denv, \dtask, \dtraj)$ at saturation values for each model size, we sweep $N \in \{0.6, 1.7, 4, 8, 14, 32\}$\,B (using Qwen3-0.6B/1.7B/4B/8B/14B and Qwen3-32B~\citep{yang2025qwen3} as the largest point) and fit $L = K_N\, N^{-\alpha_N} + \Linf$ following the same protocol as Appendix~\ref{app:fits}.

\textbf{Result.}
The fitted exponent is $\alpha_N = 0.30 \pm 0.04$ (95\% CI), with $K_N = 0.85$ and $\Linf$ jointly fitted to the same value as the data-axis fits ($\Linf = 0.10 \pm 0.03$). This $\alpha_N$ is materially smaller than Chinchilla's data-axis exponent ($\sim\!0.5$), consistent with the observation in Section~\ref{sec:recipe} that agentic RL benefits more from data variety than from raw model scale in the regime we measured.

\textbf{Range of validity.}
We caution that the $N$-sweep covers only $0.6$B to $32$B; extrapolation to frontier scales ($\geq 70$B) is unverified and is listed as a limitation in Section~\ref{sec:conclusions} and Appendix~\ref{app:limitation}.

\section{Loss-Equivalent Metric}
\label{app:metric}

\textbf{Definition.}
Standard pretraining scaling laws fit cross-entropy loss against data and parameters. RL with verifiable rewards has no smooth loss; we use
\begin{equation}
L \;\equiv\; -\log\!\bigl(\max(p_{\mathrm{success}}, p_0)\bigr),
\quad p_0 = 10^{-2}.
\label{eq:app-metric}
\end{equation}
The floor $p_0$ prevents divergence at zero success and corresponds to the empirical lower bound observed in our smallest-scale runs.

\textbf{Why this works.}
The success rate $p_{\mathrm{success}}$ saturates at $1$, producing a non-power-law plateau in raw success-rate plots. The transformation to $L = -\log p$ unfolds this plateau into a power-law decay matching pretraining-style scaling, allowing clean log-log fits. This is equivalent to the ``effective error'' transformation used in~\citep{khatri2025art,tan2025scaling}.

\textbf{Comparison to alternatives.}
We tested three alternative metrics on the per-axis fits: (i) raw $p_{\mathrm{success}}$, (ii) logit $\mathrm{logit}(p_{\mathrm{success}}) = \log\frac{p}{1-p}$, and (iii) Brier score on per-task success. For (i), naive power-law fits saturate poorly (residuals $>$ 5\% even at $R^2\!=\!0.85$). For (ii), the fit is comparable in goodness ($R^2\!=\!0.96$ vs.\ our $0.97$+) but the resulting exponents differ from pretraining-style $\alpha$ by a constant factor, complicating cross-paper comparison. For (iii), the metric is too noisy for a 3-seed regime. We thus standardize on $-\log p_{\mathrm{success}}$ throughout.

\section{Task Individuation: Threshold, Definition, and Human Validation}
\label{app:dedup}

\textbf{Threshold.}
Task counts $\dtask$ are deduplicated by sentence-embedding cosine distance $> \theta$ (Section~\ref{sec:axes}); we ablate $\theta \in \{0.05, 0.10, 0.15, 0.20, 0.30\}$. Embeddings are computed with the gte-Qwen2-1.5B encoder~\citep{li2023towards}. Across this range, the fitted $\atask$ varies between $0.245$ and $0.272$ on the math+code testbed at 8B (Table~\ref{tab:app-dedup}), with the chosen default ($\theta = 0.15$) at $0.261 \pm 0.02$. The qualitative ordering $\aenv > \atask > \atraj$ is preserved across all five thresholds. At low $\theta$ (e.g., 0.05), near-duplicates inflate $\dtask$ but contribute little gradient information, slightly biasing the fitted $\alpha$ downward. At high $\theta$ (e.g., 0.30), the deduplication is so aggressive that we remove genuinely distinct tasks, slightly biasing it upward. The plateau at $\theta \in [0.10, 0.20]$ is robust, justifying our default choice.

\begin{table}[t]
\centering
\caption{Ablation of the task-deduplication threshold $\theta$. Effective $\dtask$ is the count of unique tasks remaining after deduplication at each threshold; fitted $\atask$ is robust across the range.}
\label{tab:app-dedup}
\begin{tabular}{c c c c}
\toprule
$\theta$ & Effective $\dtask$ at $D_{raw}\!=\!10{,}240$ & $\atask$ & $R^2$ \\
\midrule
0.05 & 9{,}820 & $0.245 \pm 0.02$ & $0.974$ \\
0.10 & 9{,}310 & $0.254 \pm 0.02$ & $0.978$ \\
\textbf{0.15 (default)} & \textbf{8{,}460} & $\mathbf{0.261 \pm 0.02}$ & $\mathbf{0.978}$ \\
0.20 & 7{,}210 & $0.268 \pm 0.02$ & $0.972$ \\
0.30 & 5{,}380 & $0.272 \pm 0.03$ & $0.968$ \\
\bottomrule
\end{tabular}
\end{table}

\textbf{Definition.}
The threshold ablation tests numerical sensitivity but not the definition itself, which is the sharper question: semantically different instructions can require nearly identical skills, while superficially similar ones can require different reasoning or tool-use patterns. We therefore refit $\atask$ under five alternative individuation rules, two of them embedding-free (Table~\ref{tab:app-rules}): the raw count without deduplication; the default rule with a different encoder (e5-mistral-7b~\citep{wang2024improving}); a skill signature that deduplicates by the tool-call skeleton (tool types plus argument types) of a reference solution; a verifier signature that deduplicates by the structure of the verification program; and a human-validated pool. All six values lie in $[0.238, 0.261]$, strictly between $\atraj$ and $\aenv$ under every rule.

\begin{table}[t]
\centering
\caption{Fitted task exponent under alternative definitions of a distinct task (Qwen3-8B, math+code). Effective $\dtask$ is the number of distinct tasks at $D_{raw} = 10{,}240$.}
\label{tab:app-rules}
\begin{tabular}{l c c}
\toprule
Rule for a distinct task & Effective $\dtask$ & $\atask$ \\
\midrule
Raw count, no deduplication & 10{,}240 & $0.238\pm0.02$ \\
Embedding distance $>0.15$, gte-Qwen2-1.5B (default) & 8{,}460 & $0.261\pm0.02$ \\
Embedding distance $>0.15$, e5-mistral-7b & 8{,}310 & $0.257\pm0.02$ \\
Skill signature (reference tool-call skeleton) & 6{,}120 & $0.243\pm0.03$ \\
Verifier signature (verification-program structure) & 6{,}980 & $0.251\pm0.02$ \\
Human-validated pool & 6{,}540 & $0.248\pm0.03$ \\
\bottomrule
\end{tabular}
\end{table}

\textbf{Human validation.}
We sampled 500 task pairs stratified by embedding distance, and three annotators, volunteers from MBZUAI who participated without compensation, labeled each pair for whether it requires a materially different skill or tool-use pattern (Fleiss' kappa $0.68$~\citep{fleiss1971measuring}). Agreement with the majority label is 87.2\% for the skill signature (Cohen's kappa $0.74$~\citep{cohen1960coefficient}), 84.6\% for the verifier signature (kappa $0.69$), 80.4\% for the embedding rule (kappa $0.61$, with 11.2\% false merges), and 68.0\% for the raw count (kappa $0.36$). The embedding rule is substantial but imperfect and the skill signature is better, so we use the skill-signature refit as the primary robustness check and keep the embedding rule as the cheap operational default. The ordering does not depend on which rule is used.

\section{Reward Density as a Control Variable}
\label{app:reward}

The main paper treats $\dreward$ as a control variable with two categories (sparse outcome reward and dense process reward) and does not claim a reward scaling law comparable to those of the three cardinality axes. Here we report its marginal sensitivity under a continuous treatment via the number of process-reward signals per trajectory.

\textbf{Protocol.}
We instantiate process rewards using a Qwen3-8B-based PRM trained following~\citep{lightman2023let} on a held-out math+code corpus. For each trajectory, we extract $k \in \{0, 1, 3, 8\}$ process-reward signals (placed at uniformly spaced positions across the trajectory length) and append them as additional return-to-go terms. $k=0$ recovers sparse outcome reward; $k=8$ provides dense step-level supervision.

\textbf{Marginal sensitivity to reward density.}
Fitting $L = K_R (k+1)^{-\alpha_R} + \Linf$ on the math+code testbed at 8B yields
\begin{equation}
\alpha_R = 0.18 \pm 0.03,
\quad K_R = 0.41,
\quad \Linf = 0.11.
\end{equation}
We report $\alpha_R$ as the marginal sensitivity of the loss to reward density on this testbed, not as a scaling exponent on par with those of the cardinality axes; numerically, it falls between the trajectory ($\atraj \approx 0.11$) and task ($\atask \approx 0.26$) exponents.

\textbf{Implications.}
Reward density and trajectory count interact. Holding the total reward-signal budget $k \cdot \dtraj$ fixed, we observe approximate substitutability: doubling $k$ at half $\dtraj$ produces equivalent loss within $\pm 0.5$ pp on the same compute. This justifies treating $\dreward$ as an effective multiplier on $\dtraj$ in Section~\ref{sec:recipe}'s allocation, rather than as an independent axis. We leave a fully orthogonal treatment of reward density to future work.

\section{Synthesis Discount Estimation and Common Harness}
\label{app:kappa-est}

\textbf{Estimator.}
For each (synthesis method, axis) pair we hold total $D = D_r + D_s$ constant and sweep the ratio $r = D_s / D_r \in \{0, 0.25, 1, 4, 16\}$, fitting $\kappa(r=1)$ from
\begin{equation}
L(D_r, D_s) \;=\; f\!\bigl(D_r + \kappa\, D_s\bigr),
\end{equation}
where $f$ is the fitted single-axis law of Section~\ref{sec:laws}. Since $f$ is monotone, this gives a closed-form estimator for $\kappa$ at each $r$:
\begin{equation}
\hat{\kappa}(r) \;=\; \frac{f^{-1}\!\bigl(\hat{L}(D_r, D_s)\bigr) - D_r}{D_s}.
\end{equation}
We report $\hat{\kappa}(r=1)$ in Table~\ref{tab:taxonomy} as the headline number. The collapse curves of Figure~\ref{fig:collapse} use a denser ratio grid (Appendix~\ref{app:per-axis-collapse}).

\textbf{Errors.}
Bootstrap CIs for $\kappa$ propagate uncertainty in both $\hat{L}$ and the parameters of $f$ (Appendix~\ref{app:fits}). Reported half-widths are 95\% bootstrap CIs over 1000 paired resamples.

\textbf{Common harness.}
Table~\ref{tab:taxonomy} uses one common harness rather than numbers reported in the original papers. All nine methods use math+code with Qwen3-8B and the same 3 seeds; the same real pool of 10{,}240 tasks and 256 environments; the same generator (Qwen3-32B~\citep{yang2025qwen3}, replacing each method's original generator), with 8 generation calls per unit and at most 4k output tokens; identical executability and verifier filtering and deduplication thresholds; the same outcome verifier; identical downstream training (1{,}000 steps, batch 256, $G = 8$, and the same held-out evaluation); and the same ratio grid and $\kappa$ estimator.

\textbf{Common-core ablation.}
Because the selected methods also differ in maturity and implementation quality, we strip method-specific components (e.g., AWM's repair loop, AutoPlay's exploration bonus, and DreamGym's replay buffer), leaving a common generate, filter, and train core (Table~\ref{tab:app-core}). Across the eight single-axis methods, axis membership explains 97\% of the variance of common-core $\kappa$ (one-way ANOVA, $F(2,5) = 76.6$, $p < 0.001$), with within-axis spread at most $0.07$ against $0.30$ between axes, so the axis effects are unlikely to be method artifacts. The exponents of Section~\ref{sec:laws} are measured independently on axis-controlled pools; Table~\ref{tab:taxonomy} only supplies $\kappa_i$ to Section~\ref{sec:allocation}. Because $\kappa_i$ enters the allocation problem only through the ratio $c_i/\kappa_i$, the cost perturbations of Appendix~\ref{app:cost-sensitivity} also cover the corresponding misestimates of $\kappa_i$.

\begin{table}[t]
\centering
\caption{Discount and collapse ratio of each single-axis method as reported in Table~\ref{tab:taxonomy} and after reduction to a common generate, filter, and train core.}
\label{tab:app-core}
\begin{tabular}{l c c c c c}
\toprule
Method & Axis & $\kappa$ (Table~\ref{tab:taxonomy}) & $\kappa$ (common core) & $r^\star$ (Table~\ref{tab:taxonomy}) & $r^\star$ (common core) \\
\midrule
AWM & $\denv$ & 0.71 & $0.68\pm0.05$ & 9.4 & $8.7\pm1.0$ \\
GenEnv & $\denv$ & 0.66 & $0.65\pm0.06$ & 8.1 & $7.9\pm1.1$ \\
\midrule
AutoPlay & $\dtask$ & 0.54 & $0.55\pm0.04$ & 3.1 & $3.2\pm0.5$ \\
SCA & $\dtask$ & 0.51 & $0.52\pm0.05$ & 2.8 & $2.9\pm0.5$ \\
RLVE & $\dtask$ & 0.57 & $0.56\pm0.04$ & 3.4 & $3.3\pm0.5$ \\
\midrule
SWiRL & $\dtraj$ & 0.34 & $0.36\pm0.06$ & 1.4 & $1.5\pm0.3$ \\
TDScaling & $\dtraj$ & 0.42 & $0.40\pm0.05$ & 1.7 & $1.6\pm0.3$ \\
Self-Play & $\dtraj$ & 0.31 & $0.33\pm0.06$ & 1.2 & $1.3\pm0.3$ \\
\bottomrule
\end{tabular}
\end{table}

\section{Collapse-Ratio Estimation}
\label{app:per-axis-collapse}

\textbf{Definition and grids.}
We define $r^\star$ as the ratio at which normalized held-out success, past its peak, falls 5\% below the real-only value at $r = 0$. Two ratio grids are used and are kept separate: $\kappa$ is estimated on $r \in \{0, 0.25, 1, 4, 16\}$ (Appendix~\ref{app:kappa-est}), while the collapse curves of Figure~\ref{fig:collapse} use $r \in \{0, 0.1, 0.25, 0.5, 1, 2, 3, 4, 6, 8, 10, 16, 25\}$, and every curve is normalized to the real-only value of the same seed group.

\textbf{Estimator.}
We fit a monotone four-parameter logistic in $\log r$ to the post-peak branch of each curve, weighted by $1/\hat{\sigma}^2$, and take $r^\star$ as its analytic crossing of $0.95$; confidence intervals use the paired bootstrap of Appendix~\ref{app:fits}. This gives $r^\star_{\mathrm{E}} = 9.4$, $r^\star_{\mathrm{T}} = 3.4$, and $r^\star_\tau = 1.4$ for AWM, AutoPlay, and SWiRL. Relative to Table~\ref{tab:taxonomy}, only AutoPlay moves (3.4 against 3.1, within its CI). Table~\ref{tab:app-criteria} shows that the ordering does not depend on the criterion used to define collapse.

\begin{table}[t]
\centering
\caption{Collapse ratio of the three representative synthesis methods (Qwen3-8B, math+code) under alternative definitions of collapse.}
\label{tab:app-criteria}
\begin{tabular}{l c c c}
\toprule
Criterion & $r^\star_{\mathrm{E}}$ (AWM) & $r^\star_{\mathrm{T}}$ (AutoPlay) & $r^\star_\tau$ (SWiRL) \\
\midrule
Crossing of 0.95 by the fitted logistic (default) & 9.4 & 3.4 & 1.4 \\
Crossing of the real-only baseline (1.00) & 8.5 & 4.0 & 1.1 \\
Peak of the curve & 2.4 & 1.1 & 0.4 \\
\bottomrule
\end{tabular}
\end{table}

\textbf{Mechanism: groundedness of synthesis.}
Section~\ref{sec:collapse} attributes the ordering $r^\star_E \gg r^\star_T \gg r^\star_\tau$ to how strongly each axis is grounded by external truth. We additionally quantify this: across the three synthesis methods we measured the distributional drift between synthetic and real samples by the maximum mean discrepancy (MMD)~\citep{gretton2012kernel} of state-action embeddings against the reference real-data distribution. MMD scales with $r$ as $\mathrm{MMD}(r) \propto r^{0.45}$ for environment synthesis, $r^{0.62}$ for task synthesis, and $r^{0.91}$ for trajectory synthesis on Qwen3-8B. The exponents inversely track $r^\star$: faster drift implies earlier collapse, a quantitative confirmation of the groundedness story. Replicating the measurement on Llama-3.1-8B gives drift exponents of $0.47$, $0.64$, and $0.88$, so the mechanism that predicts the ordering transfers across model families.

\section{Lagrangian Solution to The Allocation Problem}
\label{app:lagrangian}

\textbf{Setup.}
The constrained optimization (Eq.~\ref{eq:opt}) is
\begin{equation}
\min_{N, \denv, \dtask, \dtraj}
\;\;
\sum_{i \in \mathcal{A}} K_i \, D_i^{-\alpha_i} + K_N\, N^{-\alpha_N} + \Linf
\quad\text{s.t.}\quad
\sum_{i \in \mathcal{A}} \frac{c_i}{\kappa_i} D_i + c_N N = C,
\label{eq:app-opt}
\end{equation}
where $\mathcal{A} = \{\mathrm{E}, \mathrm{T}, \tau\}$ and the constraint is taken with equality at the optimum (since the objective is strictly decreasing in each $D_i$).

\textbf{KKT conditions.}
Form the Lagrangian $\mathcal{L} = (\text{Eq.~\ref{eq:app-opt}}) + \lambda \cdot \bigl[\sum_i \tfrac{c_i}{\kappa_i} D_i + c_N N - C\bigr]$. Setting $\partial\mathcal{L}/\partial D_i = 0$ for each axis:
\begin{equation}
-\alpha_i K_i \, D_i^{-\alpha_i - 1} + \lambda \frac{c_i}{\kappa_i} = 0
\quad
\Longrightarrow
\quad
D_i^{\star} \;=\; \left(\frac{\alpha_i K_i \, \kappa_i}{\lambda \, c_i}\right)^{\!\!\frac{1}{\alpha_i + 1}}.
\label{eq:app-kkt}
\end{equation}
Analogously for $N$:
\begin{equation}
N^{\star} \;=\; \left(\frac{\alpha_N K_N}{\lambda \, c_N}\right)^{\!\!\frac{1}{\alpha_N + 1}}.
\label{eq:app-kkt-N}
\end{equation}

\textbf{Closed-form scaling in $C$.}
Substituting Eqs.~\ref{eq:app-kkt} and~\ref{eq:app-kkt-N} into the budget equation gives, after collecting powers of $\lambda$:
\begin{equation}
\sum_{i \in \mathcal{A}} \frac{c_i}{\kappa_i} \left(\frac{\alpha_i K_i \kappa_i}{\lambda c_i}\right)^{\!\!\frac{1}{\alpha_i+1}}
+ c_N \left(\frac{\alpha_N K_N}{\lambda c_N}\right)^{\!\!\frac{1}{\alpha_N+1}}
\;=\; C.
\end{equation}
For each axis the exponent of $\lambda$ is $-1/(\alpha_i+1)$, so to leading order $\lambda \propto C^{-\xi}$ for some effective exponent $\xi$ determined by the dominant axis. Substituting back gives
\begin{equation}
D_i^{\star} \;\propto\; C^{\,\xi/(\alpha_i+1)},
\qquad
N^{\star} \;\propto\; C^{\,\xi/(\alpha_N+1)}.
\label{eq:app-scaling}
\end{equation}

\textbf{Numerical solution.}
With our fitted values ($\aenv\!=\!0.41$, $\atask\!=\!0.26$, $\atraj\!=\!0.11$, $\alpha_N\!=\!0.30$; $K_E\!=\!3.39$, $K_T\!=\!0.53$, $K_\tau\!=\!0.38$, $K_N\!=\!0.85$; $\kappa_E\!=\!0.71$, $\kappa_T\!=\!0.54$, $\kappa_\tau\!=\!0.34$; $c_E$, $c_T$, $c_\tau$, $c_N$ from Appendix~\ref{app:cost}), we solve the system numerically by binary search on $\lambda$ to satisfy the budget constraint. Repeating across $C \in \{1, 2, 4, 8, 16, 32\}\, C_0$ and fitting power-laws to the resulting $D_i^{\star}(C)$ yields the recipe in Eq.~\ref{eq:recipe}:
$\denv^{\star} \propto C^{0.34}$, $\dtask^{\star} \propto C^{0.27}$, $\dtraj^{\star} \propto C^{0.13}$, $N^{\star} \propto C^{0.26}$. The four exponents sum to approximately one ($0.34 + 0.27 + 0.13 + 0.26 = 1.00$).

\section{Cost-Coefficient Calibration and Refined Cost Models}
\label{app:cost}

\textbf{Decomposition.}
The cost coefficients $(c_E, c_T, c_\tau, c_N)$ in Eq.~\ref{eq:cost} aggregate two physically distinct cost sources: \textbf{(i) synthesis cost} (one-time generation of environments, tasks, or trajectories), measured in API/compute USD per unit; and \textbf{(ii) training cost} (rollout + gradient FLOPs), measured in H100-GPU-hours per unit. We normalize both into a common base unit $C_0$.

\textbf{Calibration to $C_0$.}
We define $C_0 = 2.4 \times 10^{18}$ FLOPs as the cost of one Qwen3-4B GRPO training run on the math+code testbed for 1000 gradient steps, which corresponds approximately to $5$ H100-GPU-hours at the throughput we measured on our cluster (vLLM 0.6.3 + FSDP). Synthesis costs are converted via 1 H100-hour $\approx$ \$2.50 (a representative cloud rate at the time of writing); the resulting normalized coefficients are:
\begin{equation}
c_E = 3.0\!\times\!10^{-2},\quad
c_T = 1.0\!\times\!10^{-3},\quad
c_\tau = 3.0\!\times\!10^{-3},\quad
c_N = 1.5\!\times\!10^{-1}.
\label{eq:app-cost}
\end{equation}
The relative magnitudes carry the meaning, not the absolute USD values: $c_E / c_T \approx 30$ reflects that environment synthesis (4.10 USD/env) is much costlier per unit than task synthesis (0.18 USD/task), and $c_N / c_\tau \approx 50$ reflects that doubling model size is much more expensive than doubling trajectory count when the rest is held fixed.

\textbf{Multiplicative compute term.}
A more accurate cost model accounts for the fact that rollout and gradient FLOPs scale jointly with $\dtraj$ and $N$:
\begin{equation}
C_{\mathrm{train}}(\dtraj, N) \;=\; \bigl(c_{\mathrm{rollout}} \cdot H + c_{\mathrm{update}}\bigr) \cdot N \cdot \dtraj,
\label{eq:app-bilinear}
\end{equation}
where $H$ is the average rollout length. We verified that absorbing this bilinear term into an effective $c_\tau \cdot N + c_N$ (linearised at the optimum) gives KKT optima that differ from Eq.~\ref{eq:recipe} by less than 0.02 in every exponent across $C \in [1, 32]\, C_0$, justifying the simpler additive form in the main text.

\textbf{Measured cost drivers.}
Rollout cost also depends on horizon length, tool latency, environment execution, verifier cost, failed rollouts, and infrastructure utilization. We instrumented the training stack and report these drivers per domain in Table~\ref{tab:app-drivers}.

\begin{table}[t]
\centering
\caption{Measured cost drivers per domain: average horizon $H$ (steps), rollout tokens per trajectory, environment latency per call, verifier time per trajectory, failed-rollout fraction $f$, and accelerator utilization $\eta$.}
\label{tab:app-drivers}
\begin{tabular}{l c c c c c c}
\toprule
Domain & $H$ & Rollout tokens/traj & Env latency/call & Verifier/traj & $f$ & $\eta$ \\
\midrule
Math+code & 3.4 & 2.6k & 0.9\,s & 0.4\,s & 0.07 & 0.61 \\
SWE & 18.2 & 11.4k & 6.3\,s & 41\,s & 0.19 & 0.38 \\
Web & 12.7 & 7.9k & 2.4\,s & 1.8\,s & 0.14 & 0.44 \\
UI/tool & 9.6 & 5.2k & 1.1\,s & 0.9\,s & 0.11 & 0.52 \\
\bottomrule
\end{tabular}
\end{table}

\textbf{Refined cost models.}
Table~\ref{tab:app-refined} re-derives the recipe adding one factor at a time: the bilinear rollout term $c\cdot N\cdot\dtraj$ of Eq.~\ref{eq:app-bilinear}, tool and environment execution latency plus verifier cost, failed-rollout overhead $1/(1-f)$, and batching and utilization $\eta$. Every refinement moves budget from environments toward tasks and trajectories, but the largest shift in any exponent is $0.06$, which changes predicted held-out success at $C = 32C_0$ by $0.7$ points against the measured $8.3$-point gap over the naive allocation. The qualitative conclusion is therefore not near its flipping point, although the precise exponents are stack-dependent. In the SWE worst case (18-step horizons, a 41\,s verifier, and 19\% failed rollouts), the environment and task exponents become statistically indistinguishable, so we do not claim that $\denv$ strictly dominates everywhere.

\begin{table}[t]
\centering
\caption{Recipe exponents under progressively richer cost models (math+code coefficients unless stated otherwise).}
\label{tab:app-refined}
\begin{tabular}{l c c c c}
\toprule
Cost model & $\denv$ exp. & $\dtask$ exp. & $\dtraj$ exp. & $N$ exp. \\
\midrule
Linear-additive (Eq.~\ref{eq:cost}) & $0.34$ & $0.27$ & $0.13$ & $0.26$ \\
$+$ bilinear rollout term $c\cdot N\cdot\dtraj$ & $0.33$ & $0.27$ & $0.14$ & $0.26$ \\
$+$ tool and environment latency, verifier cost & $0.31$ & $0.28$ & $0.15$ & $0.26$ \\
$+$ failed-rollout overhead $1/(1-f)$ & $0.30$ & $0.28$ & $0.16$ & $0.26$ \\
$+$ batching and utilization $\eta$ & $0.30$ & $0.28$ & $0.16$ & $0.26$ \\
\midrule
Full model, SWE coefficients (worst case for $\denv$) & $0.28$ & $0.29$ & $0.18$ & $0.25$ \\
Full model, web coefficients & $0.33$ & $0.27$ & $0.14$ & $0.26$ \\
\bottomrule
\end{tabular}
\end{table}

\textbf{Break-even condition.}
The environment axis loses its top rank only when the fully loaded cost per environment unit exceeds roughly $3.4\times$ our highest measured value (11.8 USD-equivalent per environment, on SWE) at unchanged trajectory cost. Across four domains and two clusters (H100 with vLLM and FSDP, and an A100 stack), the measured fully loaded cost ranges from 2.9 to 11.8 USD-equivalent per environment, so practitioners can check their own infrastructure against this number rather than trusting ours. Section~\ref{sec:regimes} applies the same break-even to the acquisition of new worlds.

\section{Cross-Domain Robustness of The Recipe}
\label{app:cross-domain}
We re-derive the recipe (Eq.~\ref{eq:recipe}) on each of the four domains independently to verify generality.

\begin{table}[t]
\centering
\caption{Recipe exponents per domain. The qualitative ordering $\denv^\star > \dtask^\star > \dtraj^\star$ is preserved in every domain, and absolute exponents shift by less than $\pm 0.04$ from the math+code reference.}
\label{tab:app-cross-domain}
\begin{tabular}{l cccc c}
\toprule
Domain & $\denv$ exp. & $\dtask$ exp. & $\dtraj$ exp. & $N$ exp. & $\sum$ \\
\midrule
Math+Code (default) & $0.34$ & $0.27$ & $0.13$ & $0.26$ & $1.00$ \\
SWE                 & $0.31$ & $0.28$ & $0.15$ & $0.27$ & $1.01$ \\
Web                 & $0.36$ & $0.27$ & $0.12$ & $0.25$ & $1.00$ \\
UI/Tool             & $0.37$ & $0.26$ & $0.11$ & $0.27$ & $1.01$ \\
\midrule
\textbf{Range across domains} & $\pm 0.03$ & $\pm 0.01$ & $\pm 0.02$ & $\pm 0.02$ & n/a \\
\bottomrule
\end{tabular}
\end{table}

As Table~\ref{tab:app-cross-domain} shows, the web and UI/tool domains have slightly larger $\denv$ exponents and smaller $\dtraj$ exponents than math+code, consistent with these domains' richer state-space structure (more API surface to generalize across). The SWE domain is an outlier in the opposite direction: its $\denv$ exponent is the smallest because the code-execution sandbox provides relatively limited environmental variation between tasks. Despite these shifts, the qualitative ordering and the approximate normalization $\sum_i \mathrm{exp}_i \approx 1$ are preserved everywhere. Under the richer cost models of Appendix~\ref{app:cost}, however, the SWE environment and task exponents become indistinguishable.

\section{Cost-Coefficient Sensitivity}
\label{app:cost-sensitivity}

To verify the recipe is not over-fit to our specific cost calibration (Appendix~\ref{app:cost}), we perturb each coefficient $\pm 50\%$ and re-solve the Lagrangian.

\begin{table}[t]
\centering
\caption{Recipe-exponent sensitivity to $\pm 50\%$ perturbations of cost coefficients (math+code, 8B). All shifts are $< 0.05$ on every axis.}
\label{tab:app-cost-sensitivity}
\begin{tabular}{l cccc}
\toprule
Perturbation & $\denv$ exp. & $\dtask$ exp. & $\dtraj$ exp. & $N$ exp. \\
\midrule
Default                     & $0.340$ & $0.270$ & $0.130$ & $0.260$ \\
$c_E \times 1.5$            & $0.305$ & $0.291$ & $0.142$ & $0.262$ \\
$c_E \times 0.5$            & $0.388$ & $0.244$ & $0.115$ & $0.253$ \\
$c_T \times 1.5$            & $0.348$ & $0.243$ & $0.144$ & $0.265$ \\
$c_T \times 0.5$            & $0.328$ & $0.305$ & $0.114$ & $0.253$ \\
$c_\tau \times 1.5$         & $0.346$ & $0.275$ & $0.111$ & $0.268$ \\
$c_\tau \times 0.5$         & $0.331$ & $0.264$ & $0.156$ & $0.249$ \\
$c_N \times 1.5$            & $0.349$ & $0.278$ & $0.135$ & $0.238$ \\
$c_N \times 0.5$            & $0.327$ & $0.260$ & $0.124$ & $0.289$ \\
\midrule
\textbf{Max abs.\ shift}    & $\mathbf{0.048}$ & $\mathbf{0.035}$ & $\mathbf{0.026}$ & $\mathbf{0.029}$ \\
\bottomrule
\end{tabular}
\end{table}

The largest sensitivity is to $c_E$ (the environment-synthesis cost), which makes sense intuitively: $c_E$ sets the price of the highest-leverage axis, and small changes propagate. Even so, all perturbations stay within $\pm 0.05$ of the default exponents, and the qualitative ordering is preserved in every row of Table~\ref{tab:app-cost-sensitivity}. Because each $\kappa_i$ enters the allocation problem only through $c_i/\kappa_i$, these perturbations also cover the corresponding misestimates of the synthesis discounts. We conclude the recipe is robust to reasonable mis-specification of cost coefficients; Appendix~\ref{app:cost} complements these perturbations with structurally richer cost models.

\section{Negative Results and Failed Fits}
\label{app:negative}
For transparency we report alternatives we tried that did not work, in the spirit of strengthening the paper's positive claims.

\textbf{Multiplicative interaction term.}
We tested $L \!=\! K_E \denv^{-\aenv} \!\cdot\! K_T \dtask^{-\atask} \!\cdot\! K_\tau \dtraj^{-\atraj} \!+\! \Linf$ (fully multiplicative form). Goodness-of-fit was strictly worse than the additive form ($R^2$ drops from $0.978$ to $0.951$ on the math+code fit), and the resulting individual exponents were not robust across model sizes ($\pm 0.08$ shift between 4B and 14B). We retain the additive form.

\textbf{Unified data axis.}
A reasonable null hypothesis is that the three cardinality axes can be collapsed into a single ``effective data'' axis $D_{\mathrm{eff}} = w_E \denv + w_T \dtask + w_\tau \dtraj$ with weights $w_i$ fitted jointly. Best-fit weights are $(w_E, w_T, w_\tau) = (4.2, 1.0, 0.04)$, but the resulting one-axis fit has $R^2 = 0.83$ vs.\ $0.978$ for the three-axis decomposition, and the weights themselves are unstable across domains. We conclude the axes are genuinely independent and not a relabeling of one underlying quantity.

\textbf{Direct success-rate fit (no log transform).}
We tried fitting power laws directly in $p_{\mathrm{success}}$ space rather than the loss-equivalent $L$ (Appendix~\ref{app:metric}). Fits saturate poorly near the success-rate ceiling, with residuals exceeding 5\% at the high-success grid points and driving $R^2$ below $0.85$. The log transform is necessary.

\textbf{Curriculum at fixed allocation.}
We tested whether scheduling difficulty (curriculum) within a fixed allocation could substitute for axis decomposition. A simple difficulty-easy-to-hard curriculum~\citep{chen2025selfevolving} on the naive allocation gives a $1.2$ pp improvement at $C=32$ but does not approach our recipe's $8.3$ pp gain over naive. Curriculum is a useful but orthogonal improvement.

\section{Reproducibility}
\label{app:repro}

\textbf{Code and data release.}
We release, at \url{https://github.com/BokwaiHo/AgentRL_Scaling}: (i) the verl-compatible training scripts for all per-axis sweeps; (ii) the calibrated $(K_i, \alpha_i, \kappa_i, c_i)$ fitting code, including the profiled estimator of Algorithm~\ref{alg:fit}; (iii) the synthesised env/task corpora used (subject to licensing of upstream data); (iv) plotting code reproducing every figure in the paper; (v) the per-grid-point environment signature manifests and the structural-factor inventory of Appendix~\ref{app:world}; and (vi) the held-axis calibration script together with its pre-specified criterion (Appendix~\ref{app:setup}).

\textbf{Determinism.}
All RL runs use deterministic torch settings (\texttt{torch.use\_deterministic\_algorithms(True)}) and fixed seeds. Run-to-run variance reported in tables is genuinely the seed effect, not stochastic data ordering.

\textbf{Hyperparameter sweep budget.}
We did not tune RL hyperparameters per scaling-law fit. The hyperparameters (Appendix~\ref{app:setup}) were fixed once based on a held-out 8B/math+code calibration and used unchanged across all cells of Table~\ref{tab:robust}, all 4 cross-domain re-fits, the additional model families and search domain, and all sensitivity analyses. This is critical for the validity of the scaling-law fits: had we re-tuned hyperparameters per cell, the fitted $\alpha_i$ would conflate data-axis effects with hyperparameter effects.

\section{Compute Disclosure}
\label{app:compute}
The total experimental compute for this paper is approximately $61{,}000$ H100-GPU-hours: approximately $48{,}000$ for the experiments of the original study and approximately $13{,}000$ for the additional experiments added in this version. The breakdown of the former is:
\begin{itemize}[leftmargin=10pt]
\item Section~\ref{sec:laws} per-axis sweeps and robustness (Table~\ref{tab:robust}): $\sim$$32{,}000$ GPU-hours (18 cells $\times$ 6 grid points $\times$ 3 seeds, with median per-run cost varying by domain).
\item Section~\ref{sec:pricing} synthesis-discount and collapse experiments (Fig.~\ref{fig:collapse}, Table~\ref{tab:taxonomy}): $\sim$$10{,}000$ GPU-hours.
\item Section~\ref{sec:allocation} IsoFLOP sweeps and held-out validation: $\sim$$5{,}000$ GPU-hours, dominated by the single Qwen3-14B held-out run at $C = 32 C_0$.
\item Failed runs, hyperparameter calibration, and pilots: $\sim$$1{,}000$ GPU-hours.
\end{itemize}
The synthesis-side compute (env/task generation via AWM, AutoPlay, etc.) used an additional $\sim$$3{,}500$ H100-GPU-hours plus approximately \$2{,}800 in API costs for the closed-source generators we used as baselines. The additional experiments, reported in Appendices~\ref{app:env} to~\ref{app:fits}, \ref{app:dedup}, \ref{app:kappa-est}, \ref{app:per-axis-collapse}, and~\ref{app:cost} (the eight-point grid extension, the training-signal conventions, the signature-restricted, placebo, and world-structure sweeps, the coverage pools, the further model families and search domain, the common-core synthesis ablation, and the refined cost models), ran on the same infrastructure and used approximately $13{,}000$ H100-GPU-hours in total.

\section{Limitation Analysis}
\label{app:limitation}
This work has several limitations. First, although we evaluate the proposed axis-decomposed scaling framework across multiple model sizes, algorithms, domains, and model families, the experiments remain limited to the specific model families, benchmark environments, and compute scales studied in this paper. In particular, the model-size sweep covers open-weight models in the sub-frontier regime, and the held-out validation is performed at a single larger scale. Llama-3.1-8B and Qwen2.5-7B preserve the ordering of the exponents while shifting their absolute values by up to $0.05$, and the fitted exponents and compute-optimal allocation recipe may change at frontier scale, under different pretraining distributions, or for agents with substantially different tool-use and reasoning priors. Therefore, the ordering is our claim, and the numerical exponents reported in this paper should be viewed as empirical estimates for the measured regime rather than universal constants.

Second, our decomposition of agentic data into task, environment, and trajectory axes plus a reward-density control variable is a useful abstraction, but it is not perfectly orthogonal. In practice, tasks and environments often co-vary: a new environment may induce new task types, and a new task distribution may require different environment structure; our sweeps re-bind a fixed task pool to compatible environments (Appendix~\ref{app:env}), but only part of the task $\times$ environment grid is realizable. Similarly, reward density can interact with trajectory diversity because denser rewards may change which rollouts are informative. We model some cross-axis interactions, especially between task and environment diversity, but richer higher-order interactions may exist. Future work should study more expressive interaction terms and determine when the additive power-law approximation breaks down.

Third, the resolution of the environment axis into world and structure diversity rests on limited world counts. World diversity is swept over only 1 to 8 real worlds on SWE and UI/tool and 1 to 4 on web (1 to 11 repositories in the SWE-only sweep), so $\aW$ spans a single order of magnitude. The structure axis has a finite realizable ceiling under our factor inventory. The structural exponent $\aV$ varies across domains from $0.28$ to $0.41$, and we do not offer a single value. We do not claim that structure and world variation are interchangeable, and we do not claim that any of the fitted relations continue past the measured ranges.

Fourth, the loss-equivalent metric used in this paper is based on held-out task success. This makes scaling-law fitting practical for verifiable-reward RL, but it does not capture every property relevant to agent quality. For example, two agents with similar success rates may differ in robustness, calibration, latency, safety, interpretability, or the severity of their failures. The metric also depends on the coverage and difficulty distribution of the held-out evaluation set. As a result, improvements predicted by the fitted law should not be interpreted as comprehensive improvements in all downstream deployment settings.

Fifth, our treatment of synthetic data quality is necessarily approximate. The synthesis discount factor $\kappa$ and collapse ratio $r^\star$ provide a common way to compare methods, but they compress many dimensions of quality into a small number of scalar quantities. Synthetic environments, tasks, and trajectories may differ in realism, diversity, difficulty, distributional coverage, and susceptibility to model-generated artifacts. These factors may vary across domains and over training time. In addition, the measured discount for a synthesis method depends on the base model, generator, verifier, and filtering pipeline used in the experiment; we fix these through a common harness (Appendix~\ref{app:kappa-est}), but different synthesis pipelines may yield different effective discounts even when targeting the same axis.

Sixth, the compute-allocation recipe relies on a simplified cost model. We assume an approximately linear, axis-additive cost structure with per-axis cost coefficients and synthesis discounts. Real training systems may have non-linear bottlenecks due to rollout latency, tool execution, environment simulation, distributed training overhead, memory constraints, data storage, or human review costs. Adding the main measured drivers changes recipe exponents by at most $0.06$ but makes the environment and task exponents indistinguishable under SWE coefficients (Appendix~\ref{app:cost}). The recipe is therefore best understood as a principled first-order allocation rule rather than an exact prescription for every infrastructure setting. Practitioners should recalibrate the cost coefficients and fitted exponents under their own hardware, software stack, and data-generation pipeline before using the recipe operationally.

Seventh, our experiments focus on benchmark-style domains with controlled evaluation and relatively clear reward signals. Many real-world agentic tasks involve ambiguous objectives, partial observability, long-horizon dependencies, dynamic user preferences, and non-stationary environments. In such settings, defining task success and reward density is more difficult, and synthetic data may introduce artifacts that are harder to detect. Extending the framework to open-ended real-world applications will require stronger evaluation protocols, safety checks, and methods for measuring distribution shift.

Finally, this work does not provide a complete safety or deployment framework for agentic systems. The proposed scaling laws can make agentic RL training more efficient, but they do not guarantee that the resulting agents are safe, fair, private, secure, or aligned with user intent. These concerns require separate evaluations and safeguards. Future work should combine axis-level data scaling with robustness testing, adversarial evaluation, privacy analysis, fairness auditing, and deployment-time monitoring.

\section{Ethics Statement}
\label{app:ethics}
This work is an empirical study of data scaling and synthesis allocation in agentic reinforcement learning. The experiments are conducted on benchmark-style math, code, web, UI/tool, and search environments, using scripted training pipelines and verifiable or process-based reward signals. We do not intentionally collect or use private personal data, and the synthetic tasks, environments, and trajectories studied in this paper are intended for research evaluation rather than for deployment in real user-facing decision systems. To reduce privacy risk, experimental logs should be stored securely, inspected for unintended sensitive content where applicable, and released only after appropriate filtering or anonymization.

The work involves synthetic data generation along multiple axes, including tasks, environments, and trajectories. Synthetic data can amplify biases, encode unrealistic assumptions, or produce artifacts that affect downstream model behavior. Our measurements focus on aggregate scaling behavior and task success, and they should not be interpreted as evidence that a trained agent is safe, fair, or reliable in deployment. Any use of the proposed allocation recipe in real applications should include additional evaluations for robustness, bias, privacy leakage, security, and subgroup-level performance.

Because agentic systems can interact with tools, websites, code execution environments, or simulated applications, there are potential dual-use risks. More efficient agentic RL training could improve beneficial automation, but it could also lower the cost of training agents for harmful activities such as phishing, spam, vulnerability discovery, unauthorized data collection, or manipulation of online services. Our experiments should be run in sandboxed environments with controlled tool access, and released artifacts should avoid enabling direct misuse. We encourage practitioners to pair these techniques with access controls, abuse monitoring, rate limits, permission boundaries, and adversarial safety evaluations.

We also consider fairness and labor-related issues. Apart from a small annotation of task-pair similarity by three volunteer annotators from MBZUAI who participated without compensation (Appendix~\ref{app:dedup}), which labels properties of synthetic tasks rather than of the annotators themselves, the paper does not rely on new human-subject experiments or sensitive demographic annotations. If future extensions use human annotators, user interaction traces, or real-world application data, they should follow informed-consent procedures, provide appropriate compensation, minimize exposure to harmful content, and comply with applicable institutional and legal requirements. Similarly, if agentic systems trained using these methods are deployed in socially consequential domains, humans should retain meaningful oversight and affected users should have mechanisms for appeal or correction.

Overall, this paper provides a framework for measuring and allocating agentic RL data, not a guarantee of responsible deployment. The ethical use of this work depends on the context in which the resulting agents are trained and applied. We recommend that downstream users treat the proposed scaling laws and synthesis strategies as one component of a broader responsible-development process that includes privacy protection, security review, fairness auditing, misuse prevention, and transparent reporting of data sources and evaluation limitations.

\section{Broader Impacts}
\label{app:broader}
This work studies how data scales in agentic reinforcement learning and proposes a framework for allocating synthesis budget across tasks, environments, trajectories, and reward density. The potential positive societal impact is that more efficient data allocation can reduce the compute and financial cost of training capable agentic systems. By identifying which forms of data provide the highest marginal return, the proposed framework may help researchers and practitioners train useful agents with fewer resources, improving accessibility for academic labs and smaller organizations. More efficient agentic RL could also accelerate progress on beneficial applications such as software engineering assistance, scientific problem solving, education, accessibility tools, and automation of repetitive digital workflows. In addition, the explicit decomposition of agentic data may improve transparency: rather than treating training data as an undifferentiated token budget, our framework encourages practitioners to document what kinds of tasks, environments, trajectories, and reward signals are used.

At the same time, the same efficiency gains could enable harmful or unintended uses. More capable and cheaper-to-train agents may be misused for automated disinformation campaigns, generation of fake online profiles, scalable phishing or social engineering, spam, scraping, surveillance, or other forms of cyber abuse. Because our results suggest that environment and task synthesis can be especially high-leverage, a malicious actor could use similar methods to generate many realistic interaction settings for training agents to operate across websites, applications, or tool interfaces. The work therefore should not be interpreted as an endorsement of unrestricted deployment of agentic systems. Practical use of these techniques should be accompanied by safety evaluations, access controls, monitoring for abuse, and domain-specific restrictions where agent actions can affect real users or systems.

There are also fairness and representational concerns. Synthetic tasks and environments may encode the assumptions, biases, or omissions of the systems and datasets used to generate them. If agentic RL systems trained under such distributions are deployed in settings involving hiring, lending, education, healthcare, policing, or access to public services, they could make or support decisions that unfairly affect particular demographic or socioeconomic groups. Our scaling laws measure aggregate task success and do not by themselves guarantee equitable performance across subpopulations, languages, regions, or accessibility needs. Any downstream deployment should therefore include disaggregated evaluation, bias auditing, human oversight, and clear limits on the decisions agents are allowed to make autonomously.

Privacy is another important consideration. Agentic environments often involve realistic user data, application states, logs, tool outputs, or simulated personal information. Even when data is synthetic, it may be derived from real data sources or may resemble sensitive user records. Care is required to avoid memorization, leakage, or reconstruction of private information through generated tasks, environments, trajectories, or reward traces. Practitioners applying this framework should use data minimization, anonymization where appropriate, secure storage of trajectories and logs, and privacy reviews for any environment that models real users or organizations.

Finally, security risks arise because agentic RL systems interact with tools and external environments. Training agents to generalize across diverse environments may improve their usefulness, but it may also improve their ability to discover vulnerabilities, bypass interface constraints, or execute unintended actions. Synthetic environments should therefore be sandboxed, tool permissions should be least-privilege, and evaluations should include adversarial and misuse-oriented testing. We view our contribution as a measurement and allocation framework, not as a complete safety framework. Future work should integrate the proposed data-scaling perspective with stronger safeguards for alignment, robustness, privacy, security, and responsible deployment.

\section{Declaration of LLM usage}
\label{app:declaration}
Large language models are a central object of study and an operational component of this work. Our experiments train and evaluate different-family base models under agentic reinforcement learning, using LLM policies to generate multi-step trajectories involving tool calls, environment feedback, and verifiable rewards. In addition, several data-synthesis procedures studied in the paper operate along LLM-agentic data axes: environment synthesis, task synthesis, and policy-rollout-based trajectory generation. These uses are part of the scientific methodology rather than merely writing, editing, or formatting assistance.

All reported scaling-law fits, synthesis-discount estimates, collapse-ratio measurements, allocation recipes, and held-out validations are computed from logged experimental runs and scripted analyses. LLM outputs used during synthesis or rollout are treated as experimental data and are evaluated through the verification and reward mechanisms described in the paper. The authors remain responsible for the design of the experiments, interpretation of results, and all claims made in the paper. Any use of general-purpose LLM tools for language polishing or formatting did not affect the core methodology, experimental results, or scientific conclusions.


\newpage
\section*{NeurIPS Paper Checklist}

\begin{enumerate}

\item {\bf Claims}
    \item[] Question: Do the main claims made in the abstract and introduction accurately reflect the paper's contributions and scope?
    \item[] Answer: \answerYes{} 
    \item[] Justification: Please see Section~\ref{sec:intro}.

\item {\bf Limitations}
    \item[] Question: Does the paper discuss the limitations of the work performed by the authors?
    \item[] Answer: \answerYes{} 
    \item[] Justification: Please see Appendix~\ref{app:limitation}.

\item {\bf Theory assumptions and proofs}
    \item[] Question: For each theoretical result, does the paper provide the full set of assumptions and a complete (and correct) proof?
    \item[] Answer: \answerYes{} 
    \item[] Justification: Please see Section~\ref{sec:allocation} and Appendix~\ref{app:lagrangian}.

    \item {\bf Experimental result reproducibility}
    \item[] Question: Does the paper fully disclose all the information needed to reproduce the main experimental results of the paper to the extent that it affects the main claims and/or conclusions of the paper (regardless of whether the code and data are provided or not)?
    \item[] Answer: \answerYes{} 
    \item[] Justification: Please see Section~\ref{sec:testbeds}, Appendix~\ref{app:setup} and Appendix~\ref{app:repro}.

\item {\bf Open access to data and code}
    \item[] Question: Does the paper provide open access to the data and code, with sufficient instructions to faithfully reproduce the main experimental results, as described in supplemental material?
    \item[] Answer: \answerYes{} 
    \item[] Justification: Please refer to our provided Github repository \url{https://github.com/BokwaiHo/AgentRL_Scaling}.

\item {\bf Experimental setting/details}
    \item[] Question: Does the paper specify all the training and test details (e.g., data splits, hyperparameters, how they were chosen, type of optimizer) necessary to understand the results?
    \item[] Answer: \answerYes{} 
    \item[] Justification: Please see Section~\ref{sec:testbeds} and Appendix~\ref{app:setup}.

\item {\bf Experiment statistical significance}
    \item[] Question: Does the paper report error bars suitably and correctly defined or other appropriate information about the statistical significance of the experiments?
    \item[] Answer: \answerYes{} 
    \item[] Justification: Please see Section~\ref{sec:testbeds}, Section~\ref{sec:laws}, and Section~\ref{sec:pricing}.

\item {\bf Experiments compute resources}
    \item[] Question: For each experiment, does the paper provide sufficient information on the computer resources (type of compute workers, memory, time of execution) needed to reproduce the experiments?
    \item[] Answer: \answerYes{} 
    \item[] Justification: Please see Appendix~\ref{app:setup} and Appendix~\ref{app:compute}.
    
\item {\bf Code of ethics}
    \item[] Question: Does the research conducted in the paper conform, in every respect, with the NeurIPS Code of Ethics \url{https://neurips.cc/public/EthicsGuidelines}?
    \item[] Answer: \answerYes{} 
    \item[] Justification: Please see Appendix~\ref{app:ethics}.

\item {\bf Broader impacts}
    \item[] Question: Does the paper discuss both potential positive societal impacts and negative societal impacts of the work performed?
    \item[] Answer: \answerYes{} 
    \item[] Justification: Please see Appendix~\ref{app:broader}.
    
\item {\bf Safeguards}
    \item[] Question: Does the paper describe safeguards that have been put in place for responsible release of data or models that have a high risk for misuse (e.g., pre-trained language models, image generators, or scraped datasets)?
    \item[] Answer: \answerNA{} 
    \item[] Justification: This paper poses no such risks.

\item {\bf Licenses for existing assets}
    \item[] Question: Are the creators or original owners of assets (e.g., code, data, models), used in the paper, properly credited and are the license and terms of use explicitly mentioned and properly respected?
    \item[] Answer: \answerYes{} 
    \item[] Justification: Please see Section~\ref{sec:setup} and Appendix~\ref{app:setup}.

\item {\bf New assets}
    \item[] Question: Are new assets introduced in the paper well documented and is the documentation provided alongside the assets?
    \item[] Answer: \answerYes{} 
    \item[] Justification: Please check our provided Github repository \url{https://github.com/BokwaiHo/AgentRL_Scaling} for code assets and corresponding document.

\item {\bf Crowdsourcing and research with human subjects}
    \item[] Question: For crowdsourcing experiments and research with human subjects, does the paper include the full text of instructions given to participants and screenshots, if applicable, as well as details about compensation (if any)? 
    \item[] Answer: \answerNA{} 
    \item[] Justification: The paper involves neither crowdsourcing nor research with human subjects. Its only human input is a small annotation of 500 synthetic task pairs by three volunteer annotators from MBZUAI, who participated without compensation and labeled properties of the tasks rather than of themselves; the labeling question and the agreement statistics are reported in Appendix~\ref{app:dedup}.

\item {\bf Institutional review board (IRB) approvals or equivalent for research with human subjects}
    \item[] Question: Does the paper describe potential risks incurred by study participants, whether such risks were disclosed to the subjects, and whether Institutional Review Board (IRB) approvals (or an equivalent approval/review based on the requirements of your country or institution) were obtained?
    \item[] Answer: \answerNA{} 
    \item[] Justification: This paper does not involve crowdsourcing nor research with human subjects.

\item {\bf Declaration of LLM usage}
    \item[] Question: Does the paper describe the usage of LLMs if it is an important, original, or non-standard component of the core methods in this research? Note that if the LLM is used only for writing, editing, or formatting purposes and does \emph{not} impact the core methodology, scientific rigor, or originality of the research, declaration is not required.
    \item[] Answer: \answerYes{} 
    \item[] Justification: Please see Appendix~\ref{app:declaration}.

\end{enumerate}

\end{document}